\documentclass[sigconf]{acmart}
\AtBeginDocument{%
  }

\setcopyright{acmlicensed}
\copyrightyear{2026}
\acmYear{2026}
\acmDOI{XXXXXXX.XXXXXXX}
\acmConference[Preprint]{}{2026}{Under Review}
\acmISBN{978-1-4503-XXXX-X/2018/06}

\usepackage{amsmath}
\usepackage{mathtools}
\usepackage{amsthm}
\usepackage{enumitem}
\usepackage{multirow}

\theoremstyle{plain}
\newtheorem{theorem}{Theorem}[section]
\newtheorem{proposition}[theorem]{Proposition}
\newtheorem{claim}[theorem]{Claim}
\newtheorem{lemma}[theorem]{Lemma}
\newtheorem{corollary}[theorem]{Corollary}
\theoremstyle{definition}
\newtheorem{definition}[theorem]{Definition}
\newtheorem{assumption}[theorem]{Assumption}

\theoremstyle{remark}

\theoremstyle{plain}

\usepackage{algorithm}
\usepackage{algpseudocode}

\begin{document}

\title{Federated Learning of Nonlinear Temporal Dynamics with \\Graph Attention-based Cross-Client Interpretability}

\author{Ayse Tursucular}
\authornote{Equal contributions.}
\affiliation{%
 \institution{Georgia Institute of Technology}
 \city{Atlanta}
 \state{GA}
 \country{USA}}

\author{Ayush Mohanty}
\authornotemark[1]
\affiliation{%
 \institution{Georgia Institute of Technology}
 \city{Atlanta}
 \state{GA}
 \country{USA}}

\author{Nazal Mohamed}
\affiliation{%
 \institution{Georgia Institute of Technology}
 \city{Atlanta}
 \state{GA}
 \country{USA}}

\author{Nagi Gebraeel}
\affiliation{%
 \institution{Georgia Institute of Technology}
 \city{Atlanta}
 \state{GA}
 \country{USA}}

\renewcommand{\shortauthors}{Tursucular et al.}

\begin{abstract}

Networks of modern industrial systems are increasingly monitored by distributed sensors, where each system comprises multiple subsystems that generate high-dimensional time-series data. These subsystems are often interdependent, making it important to understand how temporal patterns observed at one subsystem relate to those observed across others. This problem is particularly challenging in decentralized settings where raw measurements cannot be shared, and client observations are heterogeneous. Moreover, in many practical deployments, each subsystem (client) operates a fixed, proprietary model that cannot be modified or retrained, limiting the applicability of existing approaches. Separately, nonlinear dynamics make cross-client temporal interdependencies inherently difficult to characterize, as they are embedded implicitly within nonlinear state-transition functions. This paper presents a vertical federated learning framework for learning one-step-ahead cross-client temporal interdependencies under these constraints. Each client maps high-dimensional local observations to low-dimensional latent states using a nonlinear state-space model, while a Graph Attention Network (GAT)-based central server learns a state-transition model over the communicated latent states. Given a candidate graph skeleton, the server learns state-dependent attention coefficients that capture cross-client interactions. We analytically relate the GAT attention coefficients to the Jacobian of the learned state-transition model, yielding a first-order characterization of cross-client temporal interdependencies. We further establish bounded closeness of the learned dynamics and Jacobians to a corresponding centralized oracle. Extensive synthetic experiments validate the proposed theoretical results on controlled systems with known graph skeletons. Experiments on real-world industrial datasets with unknown graph skeletons demonstrate superior federated root cause analysis over existing baselines.
\end{abstract}


\setcounter{page}{1}
\twocolumn

\maketitle

\section{Introduction}\label{sec:Introduction}

Modern industrial networks are increasingly monitored by distributed sensors and advanced instrumentation. In many deployments, each system consists of multiple geographically distributed subsystems \cite{DistManuf, DistDataCenters, DRLgrid} that generate high-dimensional time-series data. Despite their decentralization, these subsystems are often interdependent: localized disturbances at one subsystem can propagate across the network and affect others over time \cite{Cascading1, Cascading2}. Understanding such temporal interdependencies is therefore critical for system monitoring, diagnostics, and reliable operation.

In practice, these systems exhibit strongly nonlinear dynamics, making cross-subsystem temporal interdependencies vary with the system’s operating state and thus difficult to interpret. While neural network models can approximate such dynamics, they typically lack structured mechanisms for exposing subsystem-level interactions. Graph-based models, particularly attention-based architectures such as the \textit{Graph Attention Network} (GAT) \cite{GATv1, GATv2}, provide a natural way to represent directed interactions while retaining expressive power for nonlinear dynamics.

However, centralized learning of such GAT models is often impractical in industrial settings. Raw observations (data) are rarely available for centralized processing due to privacy, communication, and data-sovereignty constraints. Moreover, subsystems frequently operate fixed proprietary models supplied by original equipment manufacturers that cannot be modified or retrained \cite{NIST_OT, NIST_ICS}. 
These challenges motivate a decentralized paradigm tailored to nonlinear dynamical systems with fixed subsystem (client) models. While clients cannot share raw observations, they can often communicate low-dimensional latent representations of local system states, raising a fundamental question: \textit{can system-wide temporal interdependencies be learned from decentralized latent representations?}

\textbf{Main Contribution.} In this work, we answer this question by proposing a \textit{vertical federated learning} (VFL) \cite{FL_review1, FL_review2} framework for learning cross-client temporal interdependencies in nonlinear dynamical systems. We consider industrial settings in which clients observe distinct, temporally aligned components of a shared dynamical system and operate fixed proprietary state estimators. Each estimator maps high-dimensional local observations to a low-dimensional latent state, allowing clients to communicate these latent states without sharing their raw observations. We model the evolution of the joint latent state across clients using a nonlinear state-space model with first-order Markov dynamics. This ensures that the next latent state depends on the system history only through the current latent state. Accordingly, we focus only on directed one-step-ahead cross-client interdependencies, characterizing how the latent state of client $n$ at time $t-1$ (denoted by $h_{t-1}^n$) contributes to the latent state of client $m$ at time $t$ (i.e., $h_{t}^m$).

A GAT-based central server learns a state-transition model from the communicated latent states. As in standard GAT message-passing formulations \cite{GATv1, GATv2}, we assume that a fixed candidate graph skeleton specifying the allowable interactions is provided \textit{a priori}. Consequently, the framework learns state-dependent attention coefficients over the candidate edges without performing graph-structure discovery. We show, both theoretically and empirically, that the attention coefficients and the first-order characterization of the dynamics via the Jacobian provide complementary yet related information about cross-client interactions. While the attention coefficients describe how strongly the model weights each neighboring client, the Jacobian \(J_{mn}(t):=\partial h_t^m/\partial h_{t-1}^n\) measures the sensitivity of the predicted next state of client \(m\) to a small perturbation in the previous latent state of client \(n\). Our contribution lies in analytically relating the GAT attention coefficients to the Jacobian blocks of the learned server-side state-transition model. To the best of our knowledge, \textbf{this is the first federated framework to establish a Jacobian-based interpretation of graph-attention coefficients in nonlinear dynamical systems.} The key technical contributions of this paper are as follows:
\begin{itemize}
    \item We formalize federated learning of cross-client temporal interdependencies in nonlinear state-space systems with fixed proprietary client models.

    \item We develop a GAT-parameterized server-side state-transition model over communicated latent states to capture cross-client dynamical interactions.

    \item We derive an analytical relationship between graph-attention coefficients and Jacobian blocks, enabling a state-dependent interpretation of learned temporal interdependencies.

    \item We establish bounded closeness of the learned dynamics and Jacobians to a corresponding centralized oracle with access to all raw client data, under the same model class.

\item We validate the theoretical analysis, including bounded closeness and interpretability, on controlled synthetic systems with known graph skeletons.

\item We demonstrate superior federated root cause analysis on real-world industrial datasets with unknown graph skeletons, outperforming existing federated learning baselines.
\end{itemize}

\section{Related Work}
\label{sec:related_work}

Centralized graph neural networks (GNNs) are a standard tool for modeling interdependencies via message passing, ranging from Interaction Networks \citep{battaglia2016interaction} to attention-based architectures such as GATs \citep{velickovic2018gat}. Spatiotemporal GNNs combine graph propagation with temporal models for forecasting \citep{li2018dcrnn,yu2018stgcn}, while adaptive graph learning jointly infers latent interaction graphs and predicts future trajectories \citep{wu2019graphwavenet,wu2020mtgnn,shang2021gts,kipf2018nri,graber2020dnri}. Similar to many GNN formulations, these methods operate on a given graph topology and only learn edge-specific message-passing weights. Importantly, these work motivate GNNs as expressive models for learning temporal interdependencies, but all assume centralized access to system-wide observations.

Most federated GNN work studies \emph{horizontal} partitions (clients own disjoint node/edge sets or subgraphs) and focuses on standard prediction tasks under non-IID graph distributions, often using FedAvg-style aggregation and benchmark suites \citep{he2021fedgraphnn}. This is often referred to as \textit{horizontal federated learning} (HFL). Because message passing couples neighbors across partitions, they handle missing cross-client neighborhoods via server-assisted aggregation \citep{zhang2021subgraphfl,zheng2021asfgnn, wu2022fedpergnn}, without explicitly modeling cross-client interactions. In contrast, \text{vertical federated learning} (VFL) allows parties to split features (or sometimes edges) and coordinate secure computation for GNN layers \citep{chen2022vfg,mai2023verfedgnn} residing at the client-level. Furthermore, these works primarily emphasize privacy mechanisms and typically assume static graph data. A closely related line of work considers VFL by aligning \textit{locally pretrained} client representations via a server-side consensus graph \citep{ma2023federated}. However, this approach does
not model any client-side temporal dynamics. Our setting differs fundamentally: \textbf{(i)} clients observe different feature views of the same dynamical system (VFL), \textbf{(ii)} the GNN resides entirely at the server while clients communicate compact latent states rather than jointly executing GNN layers, and \textbf{(iii)} the objective is learning nonlinear temporal interdependencies under communication constraints rather than privacy preservation or static graph prediction.

A related line of work studies causal discovery by learning directed acyclic graphs (DAGs) from observed data using either continuous optimization \citep{zheng2018notears,yu2019daggnn,lachapelle2020grandag,pamfil2020dynotears} or conditional independence tests \citep{kalisch2007estimating, runge2019detecting, runge2020discovering}. Federated extensions adapt these formulations to distributed datasets while retaining the objective of recovering a causal graph \citep{ng2022fedbn,gao2023feddag,FDBNL, FedPC, FedCSL}. Furthermore, these methods are developed in HFL settings and do not explicitly model cross-client interactions. In contrast, our goal is not causal graph discovery. Instead, we assume a given candidate graph skeleton and learn state-dependent cross-client interdependencies over its edges.


\section{Preliminaries}
\subsection{Nonlinear State-Space Modeling}

We consider a nonlinear state-space model (SSM) with a latent state
$h_t \in \mathbb{R}^p$ and observation
$y_t \in \mathbb{R}^d$. The dynamical system then evolves according to
\begin{equation}\label{eq:nonlinearStatespace}
h_t = f(h_{t-1}) + w_t,
\qquad
y_t = g(h_t) + v_t,
\end{equation}
where $f(\cdot)$ is a nonlinear \textit{state-transition function},
$g(\cdot)$ is a nonlinear \textit{observation function},
and $w_t$, $v_t$ denote process and observation noise, respectively.

\subsection{State Estimation}

When the SSM is nonlinear, recursive estimation of the latent
state can be performed using an Extended Kalman Filter (EKF).
The EKF proceeds by alternating between a \emph{prediction} step and an
\emph{estimation} (correction) step. 

In the prediction step, the previous estimated state $\hat h_{t-1}$
is propagated forward through the nonlinear dynamics to obtain a predicted state $\tilde{h}_t$ such that, 
\begin{equation}
\tilde{h}_t = f(\hat h_{t-1}),
\end{equation}
where $\tilde{h}_t$ represents the belief about the current state before observing $y_t$. In the estimation step, this prediction is updated using the new observation $y_t$ as follows: 
\begin{equation}
\hat h_t = \tilde{h}_t + K_t\big(y_t - g(\tilde{h}_t)\big),
\end{equation}
where $K_t$ is the Kalman gain that determines how the prediction $\tilde{h}_t$ is corrected based on the
discrepancy between the observed measurement $y_t$ and its predicted value $g(\tilde{h}_t)$.



\subsection{Jacobian for Temporal Interdependencies}

To characterize directed temporal interdependencies in nonlinear dynamical systems, we decompose the latent state as \(h_t := \big[(h_t^1)^\top,\dots,(h_t^M)^\top\big]^\top\), consider its evolution under the nonlinear transition \(h_t=f(h_{t-1})\), and define the corresponding Jacobian \(J_t:=\partial f(h_{t-1})/\partial h_{t-1}\). The Jacobian is block structured, where the off-diagonal block
\begin{equation}
J_{ij}(t):=\frac{\partial h_t^i}{\partial h_{t-1}^j}, \qquad i\neq j,
\end{equation}
captures how perturbations in component \(j\) at time \(t-1\) influence component \(i\) at time \(t\). Consequently, the collection of off-diagonal blocks characterizes directed temporal interdependencies between distinct components of the latent state.

\subsection{Graph Attention Networks}

Graph Attention Networks (GATs) model interdependencies among latent state components by
combining a given graph skeleton with attention-based message passing.
The graph specifies which components may interact, while attention mechanisms
adaptively weight these interactions based on the current state.

Let $A \in \{0,1\}^{M \times M}$ denote an adjacency matrix known \textit{a priori}, and provided as input
to the GAT, which encodes the graph skeleton.
The neighborhood of each latent state component, denoted by $\mathcal N(.)$, is inferred
directly from $A$, and determines which components participate in message passing. Given latent states $h_{t-1}^i \in \mathbb{R}^p$, a GAT layer updates each node by
aggregating information from its neighbors $j \in \mathcal N(i)$.
Each node state is first linearly transformed,
\begin{equation}\label{GAT_1}
z_{t-1}^i = W h_{t-1}^i,
\end{equation}
where $W$ is a learnable weight matrix.
For each $j \in \mathcal N(i)$, an unnormalized attention score
(\emph{edge score}) is given by,
\begin{equation}\label{GAT_2}
e_{ij} = a^\top \!\left[ z_{t-1}^i \,\|\, z_{t-1}^j \right],
\end{equation}
where $a$ is a learnable vector and $[\cdot \| \cdot]$ denotes concatenation.
The attention scores $e_{ij}$ are then normalized across the neighborhood of node $i$ using a
softmax to obtain attention coefficients $\alpha_{ij}$ as, 
\begin{equation}\label{GAT_3}
\alpha_{ij} =
\frac{\exp(e_{ij})}
{\sum_{k \in \mathcal N(i)} \exp(e_{ik})},
\end{equation}
ensuring that $\sum_{j \in \mathcal N(i)} \alpha_{ij} = 1$.
The resulting attention coefficients $\alpha_{ij}$ represent the relative contributions of
neighboring state components. With $\sigma(\cdot)$ as a nonlinear activation function, the updated state is then obtained via attention-weighted aggregation as, 
\begin{equation}
h_t^i =
\sigma\!\left(
\sum_{j \in \mathcal N(i)} \alpha_{ij}\, z_{t-1}^j
\right),
\end{equation}

\section{Problem Setting}\label{sec:ProblemSetting}
\begin{figure*}[t]
    \centering
    \includegraphics[width=\textwidth]{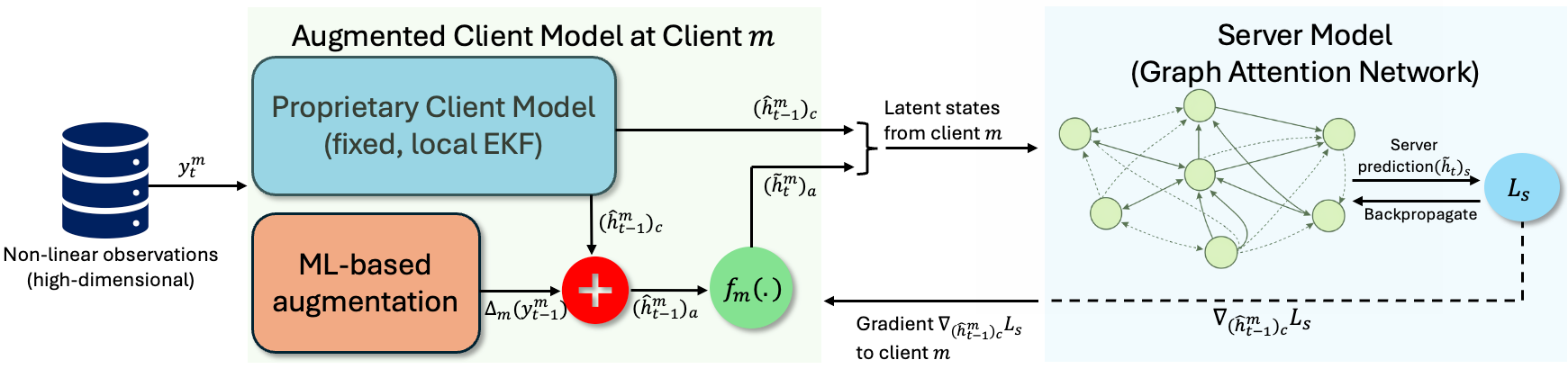}
    \caption{Schematic of our proposed framework with communication between client $m$ and the server}
    \label{fig:flowchart}
\end{figure*}

We consider a federated nonlinear dynamical system with $M$ clients and a central server (Fig.~\ref{fig:flowchart}). At time $t$, client $m\in\{1,\ldots,M\}$ observes a high-dimensional observations $y_t^m\in\mathbb{R}^{d_m}$ generated from a low-dimensional latent state $h_t^m\in\mathbb{R}^{p_m}$, where $p_m\ll d_m$. Clients operate only on local observations and never share those raw data (observations). Instead, each client $m$ employs a proprietary EKF that uses local data $y_m^t$ to produce the predicted state $(\tilde h_t^m)_c$ and estimated state $(\hat h_t^m)_c$. This introduces several challenges in learning and interpreting temporal interdependencies, mentioned next.\\

\noindent \textbullet{} \textbf{\underline{Challenge 1}.}
The proprietary EKF estimates ${(\tilde h_t^m)}_c$ and ${(\hat h_t^m)}_c$ model only local dynamics and therefore cannot capture cross-client temporal interdependencies.

\noindent - \textbf{Solution (\textit{Server model}).}
The server learns a graph-attention state-transition model over the client latent states,
$\{(\hat h_t^m)_c\}_{m=1}^M$.
We restrict this transition to admit a graphical representation
parameterized by a GAT, producing server-side state
predictions ${(\tilde h_t^m)}_s$:
\begin{equation}
\label{eq:serverstates}
{(\tilde h_t^m)}_s =
\sigma\!\left(
\sum_{n \in \mathcal N(m)}
\alpha_{mn}\, W_{mn}\, {(\hat h_{t-1}^n)}_c
\right),
\end{equation}
In Eq. \eqref{eq:serverstates}, $\alpha_{mn}$s are the attention coefficients and $W_{mn}$s are the learnable
transformations. The use of GAT \textbf{allows the server to capture directed temporal interdependencies
through an explicit graph structure,} instead of any unstructured model. 
\\

\noindent \textbullet{} \textbf{\underline{Challenge 2}.}
The server states ${(\tilde h_t^m)}_s$ cannot be directly shared with clients, creating a gap between globally learned interdependencies and locally available state estimates (from proprietary EKF).

\noindent - \textbf{Solution (\textit{Client model}).}
Each client augments its proprietary EKF state using a learnable nonlinear
mapping.
Given the estimated state $(\hat h_t^m)_c$ from the proprietary model, the
augmented state is, 
\begin{equation}\label{eq:augmentation}
(\hat h_t^m)_a = (\hat h_t^m)_c + \Delta_m\!\left(y_{t-1}^m; \theta_m\right),
\end{equation}
where $\Delta_m(\cdot)$ denotes a learnable augmentation function parametrized by $\theta_m$, and ${(\hat h_t^m)}_a$ is the estimated state by the augmented model.
\textbf{Note that the augmentation step leaves the proprietary EKF's states unchanged.} The augmented state evolves as shown in Eq. \eqref{augmented_evolve} to produce ${(\tilde h_t^m)}_a$, the predicted state of the augmented model. In Eq. \eqref{augmented_evolve}, $f_m(.)$ is client $m$'s state-transition function. 
\begin{equation}\label{augmented_evolve}
{(\tilde h_t^m)}_a = f_m\!\left({(\hat h_{t-1}^m)}_a\right),
\end{equation}

\noindent \textbullet{} \textbf{\underline{Challenge 3}.}
Without direct access to server states ${(\tilde h_t^n)}_s$ or to raw observations
$\{y_t^n\}_{n \neq m}$ from other clients, the augmented model cannot independently learn the global temporal interdependencies.

\noindent - \textbf{Solution (\textit{Communication Protocol}).} The server minimizes a
loss $L_s$ and communicates the gradient
$\nabla_{(\tilde h_t^m)_a} L_s$ to each client $m$ .
The clients then incorporate this gradient to update the parameters of their
augmentation function $\Delta_m(\cdot)$, enabling the augmented states to align
with the globally learned temporal interdependencies. \textbf{This iterative optimization aligns the client and server latent states over several communication rounds.} 
\\

\noindent \textbullet{} \textbf{\underline{Challenge 4}.}
The learned latent states encode temporal interdependencies implicitly in the model parameters, without a principled interpretation of which clients are influencing the others.


\noindent - \textbf{Solution (\textit{Interpretability}).} Section~\ref{sec:InterpretingJacobian} develops a Jacobian-based interpretation of the learned dynamics, while Section~\ref{sec:Convergence} establishes its bounded closeness to a centralized oracle. Combined with the GAT's attention coefficients, the Jacobian provides an interpretable characterization of temporal interdependencies.

\section{Methodology}\label{sec:Methodology}

\subsection{Client-Side Modeling}

\begin{assumption}[\textbf{Proprietary Model}]
Each client $m$ maintains a proprietary EKF-based state estimator that produces
the predicted state ${(\tilde h_t^m)}_c$ and corrected state estimate
${(\hat h_t^m)}_c$ using only local observations $y_t^m$. \textbf{The internal dynamics
of this model are fixed and cannot be modified.} 
\end{assumption}

\noindent \textbf{Client Loss.} Following the client augmentation model introduced in
Eq.~\eqref{eq:augmentation} and its state-transition dynamics in
Eq.~\eqref{augmented_evolve}, each client minimizes the local reconstruction
loss
\begin{equation}
(L_m)_a
=
\left\|
y_t^m-g_m\!\left({(\tilde h_t^m)}_a\right)
\right\|_2^2,
\end{equation}
where $g_m(\cdot):\mathbb{R}^{p_m}\rightarrow\mathbb{R}^{d_m}$ denotes client $m$'s observation function.

The augmentation function's parameters $\theta_m$ are updated using both the client loss
and the server loss (Section \ref{sec:server}). The server-induced gradient is computed via the chain
rule,
\begin{equation}
\nabla_{\theta_m}L_s
=
\nabla_{{(\tilde h_t^m)}_a}L_s\;
\nabla_{\theta_m}{(\tilde h_t^m)}_a,
\end{equation}
where $\nabla_{{(\tilde h_t^m)}_a}L_s$ is received from the server (Section \ref{sec:server}). The
resulting parameter update is
\begin{equation}
\theta_m^{k+1}
=
\theta_m^k
-\eta_1\nabla_{\theta_m^k}(L_m)_a
-\eta_2\nabla_{\theta_m^k}L_s.
\end{equation}

\noindent \textbf{Client-to-Server Communication.}
At each communication round, client $m$ transmits the tuple
$\left((\hat h_t^m)_c,(\tilde h_t^m)_a\right)$ to the server.




\subsection{Server-Side Graph Modeling}\label{sec:server}

\begin{assumption}[\textbf{Privacy Constraint}]
The server neither has access to raw observations nor can it transmit latent
states to the clients.
\end{assumption}

\noindent \textbf{Server Loss.} Following the server-side GAT model introduced in
Eq.~\eqref{eq:serverstates}, the server learns the graph-attention parameters
by minimizing the loss
\begin{equation}\label{eq:serverloss}
L_s
=
\frac{1}{T}\sum_{t=1}^{T}\sum_{m=1}^{M}
\left\|
{(\tilde h_t^m)}_s-
{(\tilde h_t^m)}_a
\right\|_2^2.
\end{equation}

\noindent \textbf{Server-to-Client Communication.}
To update the client augmentation models without violating federated
constraints, the server transmits the gradient
$\nabla_{{(\tilde h_t^m)}_a}L_s$ to client $m$.

\noindent \textbf{Note.}
The use of a GAT assumes that the graph skeleton (adjacency matrix) is known
\emph{a priori}. This is reasonable in engineered systems where domain
knowledge specifies the interconnections, while the edge strengths are learned
decentrally.

\subsection{Method Properties}
In this section, we formally list the key properties of the proposed framework, showing how they collectively address the
four challenges outlined in Section~\ref{sec:ProblemSetting}.

\begin{claim}[\textbf{Implicit Encoding of Interdependencies}]\label{claim1}
Each client-side augmentation model $\Delta_m(\cdot)$ implicitly encodes
cross-client temporal interdependencies, despite operating only on local
observations $y_t^m$ and local EKF state estimates $(\hat h_t^m)_c$.
\end{claim}

This property addresses \textbf{Challenges 1 and 3}.
While the proprietary EKF estimates $(\hat h_t^m)_c$ depend solely on local
observations and cannot capture cross-client effects, the augmentation model
$\Delta_m(\cdot)$ is trained using gradients
$\nabla_{(\tilde h_t^m)_a} L_s$ communicated by the server.
As a result, although evaluated only on local inputs, 
$\Delta_m(\cdot)$ is shaped by the global cross-client structure learned at the
server.

\begin{claim}[\textbf{Alignment of Client and Server States}]\label{claim2}
The augmented client states $(\tilde h_t^m)_a$ and the server-predicted states
$(\tilde h_t^m)_s$ are aligned through iterative optimization. 
\end{claim}

In the presence of cross-client temporal interdependencies, the proprietary
client states $(\tilde h_t^m)_c$ and the server predictions $(\tilde h_t^m)_s$
generally differ, since the former ignore cross-client effects while the latter
explicitly model them.
The server loss in Eq.~\eqref{eq:serverloss} enforces consistency between
$(\tilde h_t^m)_s$ and $(\tilde h_t^m)_a$, and its gradients are propagated back to
the clients to update their augmentation models.
This bidirectional coupling aligns client-side and server-side state
representations without sharing raw observations or server states, thereby
addressing \textbf{Challenge 2}.

\begin{claim}[\textbf{Interpretable Temporal Interdependencies}]\label{claim3}
The server learns a graph-structured nonlinear state-transition model in which
attention coefficients $\alpha_{mn}(t)$ regulate the aggregation of client states,
and directed temporal interdependencies are characterized by the Jacobian
$J_{mn}(t)
=
\frac{\partial (\tilde h_t^m)_s}{\partial (\hat h_{t-1}^n)_c}$.
\end{claim}

This claim addresses \textbf{Challenge 4}. While the attention coefficients
$\alpha_{mn}(t)$ quantify the relative weighting assigned to neighboring client
states in the server-side GAT, the Jacobian
$J_{mn}(t)=\frac{\partial (\tilde h_t^m)_s}{\partial (\hat h_{t-1}^n)_c}$
characterizes the resulting temporal influence. A detailed discussion on these quantities, including their analytical relationship, is
provided in Section~\ref{sec:InterpretingJacobian}.



\section{Interpreting Temporal Interdependencies}\label{sec:InterpretingJacobian}
Two
distinct but complementary questions arise when analyzing cross-client temporal
interdependencies:\\
\noindent -- \textbf{Q1.}
\emph{Given the graph topology, how
strongly is each neighbor $n \in \mathcal{N}(m)$ weighted when predicting client
$m$’s future latent state?} \\
\noindent -- \textbf{Q2.}
\emph{How does a perturbation in client $n$’s past latent state affect client
$m$’s future latent state under the learned nonlinear dynamics?}

The question \textbf{Q1.} is addressed by the attention coefficients $\alpha_{mn}$, which
assign a state-dependent relative weight to each neighbor’s
message in the state-transition mechanism at the server. Whereas \textbf{Q2.} is addressed by the Jacobian block
$
J_{mn}(t)
\;:=\;
\frac{\partial (\tilde h_t^m)_s}{\partial (\hat h_{t-1}^n)_c},
$
which provides a state-dependent measure of temporal influence. \footnote{The use of the Jacobian follows naturally from nonlinear state-space theory. In
linear SSMs, temporal interdependencies are explicitly encoded by
the state-transition matrix. In nonlinear SSMs, this notion
generalizes to the Jacobian of the state-transition function, which
characterizes how each component of the future state locally depends on past
state variables. Consequently, the Jacobian provides a state-dependent,
directional characterization of temporal interdependencies in nonlinear
dynamical systems.}

While $\alpha_{mn}(t)$ and $J_{mn}(t)$ are related, \textbf{they capture distinct aspects
of cross-client temporal interdependencies.} The attention coefficients
$\alpha_{mn}(t)$ encode the relative structural weighting of client $n$’s state
in the server’s prediction of client $m$, as determined by the graph and the
attention mechanism. In contrast, the Jacobian $J_{mn}(t)$ captures the resulting
dynamical influence after accounting for nonlinear state evolution. \textbf{Notably, the
activation function $\sigma(\cdot)$ does not appear explicitly in
$\alpha_{mn}(t)$, but enters the Jacobian through $\sigma'(\cdot)$, allowing
nonlinear effects to modulate temporal
influence.} A closed-form expression of $J_{mn}(t)$ in terms of $\alpha_{mn}(t)$ is given in Proposition \ref{prop:jacobian_gat} below. 

\begin{proposition}
\label{prop:jacobian_gat}
The Jacobian $J_{mn}(t)$ and the attention coefficient $\alpha_{mn}(t)$ are related as
\begin{equation}
\label{eq:jacobian_alpha}
\begin{split}
J_{mn}(t)
&=
\operatorname{diag}\!\big(\sigma'(s_m(t))\big)
\Bigg[
\alpha_{mn}(t)\, W_{mn}
+
\sum_{r \in \mathcal{N}(m)}
W_{mr}\\
&
\times(\hat h_{t-1}^r)_c\ 
\alpha_{mr}(t)
\big(\delta_{rn} - \alpha_{mn}(t)\big)
\frac{\partial e_{mn}(t)}{\partial (\hat h_{t-1}^n)_c}
\Bigg]
\end{split}
\end{equation}
where $s_m(t):=
\sum_{r \in \mathcal{N}(m)}
\alpha_{mr}(t)\, W_{mr}\, (\hat h_{t-1}^r)_c$, and 
$
\delta_{rn}
$ is the Kronecker delta. 
\end{proposition}

The expression in Eq.~\eqref{eq:jacobian_alpha} highlights two pathways through
which client $n$ can influence client $m$ over time: \textbf{(i)} a direct pathway
proportional to $\alpha_{mn}(t) W_{mn}$, and \textbf{(ii)} an indirect pathway
arising from the dependence of the attention mechanism on the input states
as discussed in Eqs. \eqref{GAT_1}-\eqref{GAT_3}. In Section~\ref{sec:Experiments}, we also see that
$\alpha_{mn}(t)$ and $\|J_{mn}(t)\|$ are often empirically correlated
because attention controls the strength of information that enters the state-transition.

\section{Bounded Closeness to a Centralized Oracle}\label{sec:Convergence}
In this section, we show that the temporal interdependencies captured by the proposed framework remain within a bounded distance of those of a centralized oracle.

\begin{definition}[\textbf{Centralized Oracle}]
The centralized oracle is a centralized EKF that aggregates observations from
all $M$ clients, denoted by $y_o^t$, to produce the predicted and estimated
states $\tilde{h}_o^t$ and $\hat{h}_o^t$, respectively:
\[
\tilde{h}_o^t = f(\hat{h}_o^{t-1}),
\qquad
\hat{h}_o^t = \tilde{h}_o^t + K^o\big(y_o^t - g(\tilde{h}_o^t)\big),
\]
where $f(.)$ and $g(.)$ denote the nonlinear state-transition and measurement
functions, respectively, and $K^o$ is the optimal Kalman gain of the oracle.
Since the vectors $y_o^t$, $\tilde{h}_o^t$, and $\hat{h}_o^t$, as well as the
matrix $K^o$, contain information across all $M$ clients, we introduce an
operator $\mathrm{Extract}_m(\cdot)$ to isolate the components of a vector or
the corresponding submatrix associated with client $m$. 
\end{definition}

In this paper, we let the $f$ of the centralized oracle admit a GAT representation. For a meaningful comparison, we
assume that the oracle and the server employ the
same GAT architecture and activation function $\sigma$. 
Let client $m$ have $N_m$ neighbors and latent state
dimension $p_m$. Consistent with Eq.~\eqref{eq:serverstates}, the oracle and
the server first transform the latent states of the neighboring clients
through their learnable transformations $W_{mn}^{o}$ and $W_{mn}^{s}$,
\begin{equation}
z_{mn}^{o} := W_{mn}^{o}\,(\hat{h}_o^{t-1})_n,
\qquad
z_{mn}^{s} := W_{mn}^{s}\,(\hat{h}_c^{t-1})_n,
\label{eq:transformed}
\end{equation}
and then, at any time $t$, update the predicted state of client
$m$ via GAT-based transitions as, 
\begin{equation}
  (\tilde{h}_o^t)_m
  = \sigma\!\Big(\sum_{n=1}^{N_m} \alpha_{mn}^o\, z_{mn}^{o}\Big)\hspace{0.1cm};
  (\tilde{h}_s^t)_m
  = \sigma\!\Big(\sum_{n=1}^{N_m} \alpha_{mn}^s\, z_{mn}^{s}\Big)
  \label{eq:update}
\end{equation}
where $\alpha_{mn}^o,\alpha_{mn}^s \in \mathbb{R}$ denote scalar attention
coefficients, $(\hat{h}_o^{t-1})_n$, $(\hat{h}_c^{t-1})_n$ denote the latent
states of neighboring clients, and $\sigma:\mathbb{R}^{p_m}\to\mathbb{R}^{p_m}$
is the activation function.\\

\noindent \textbf{Notation.} Define the transformed neighbor representation matrices
\[
Z_{o,m}^{t-1}
:=
\big[\, z_{m1}^{o} \; z_{m2}^{o} \; \cdots \; z_{mN_m}^{o} \,\big]
\in \mathbb{R}^{p_m\times N_m},
\] 
\[
Z_{s,m}^{t-1}
:=
\big[\, z_{m1}^{s} \; z_{m2}^{s} \; \cdots \; z_{mN_m}^{s} \,\big]
\in \mathbb{R}^{p_m\times N_m}.
\]
Let the oracle and server attention vectors be
\[
\alpha_m^o := (\alpha_{m1}^o,\dots,\alpha_{mN_m}^o)^\top,
\hspace{0.1cm} \text{and} \hspace{0.1cm}
\alpha_m^s := (\alpha_{m1}^s,\dots,\alpha_{mN_m}^s)^\top,
\]
Then,
$
(\tilde{h}_o^t)_m = \sigma(Z_{o,m}^{t-1} \alpha_m^o),
\hspace{0.1cm} \text{and} \hspace{0.1cm}
(\tilde{h}_s^t)_m = \sigma(Z_{s,m}^{t-1} \alpha_m^s).
$

We assume that the EKF dynamics used by both the centralized oracle and the
proprietary models (at the clients) are stable and convergent. Then the
following result holds:

\begin{lemma}[\textbf{State boundedness}]
\label{lemma:stateboundedness}
For sufficiently large $t$, there exist constants $\varepsilon_1,\varepsilon_2
\ge 0$ such that, for all clients $m$ and all neighbors $n$,
\[
\|(\tilde{h}_s^t)_m - (\tilde{h}_o^t)_m\| \le \varepsilon_1,
\qquad
\|z_{mn}^{s} - z_{mn}^{o}\| \le \varepsilon_2,
\]
and consequently $\|Z_{s,m}^{t-1} - Z_{o,m}^{t-1}\|_F \le \sqrt{N_m}\,\varepsilon_2$.
\end{lemma}

Lemma~\ref{lemma:stateboundedness} establishes that the states predicted by both
the server-side GAT and the centralized oracle remain bounded. However, bounded
states alone do not guarantee that the two models induce comparable temporal
interdependencies. This reflects the well-known identifiability issue in
nonlinear state-space models, where distinct latent dynamics can produce
indistinguishable state trajectories. To ensure identifiability, we impose an additional regularity condition on the activation function.
\begin{assumption}[\textbf{Bi-Lipschitz activation}]
The activation function $\sigma$ is monotone and bi-Lipschitz on the operating
range; that is, there exist constants $L_\sigma,L_{\sigma^{-1}}>0$ such that
\[
\|\sigma(u)-\sigma(v)\| \le L_\sigma \|u-v\|;
\hspace{0.05cm}
\|u-v\| \le L_{\sigma^{-1}} \|\sigma(u)-\sigma(v)\|.
\]
\end{assumption}

\begin{theorem}[\textbf{Bounded attention}]
\label{thm:param_conv}
Under the preceding assumptions, there exists a constant $C_h>0$ such that the
difference between the attention coefficients learned by the oracle and the
server-side GAT, projected onto the row space of $Z_{s,m}^{t-1}$, is bounded as
\[
\big\|P_{\operatorname{row}(Z_{s,m}^{t-1})}\big(\alpha_m^o - \alpha_m^s\big)\big\|
\;\le\;
\frac{C_h\,\varepsilon_2 + L_{\sigma^{-1}}\,\varepsilon_1}
     {\sigma_{\min}^{+}(Z_{s,m}^{t-1})},
\]
where $P_{\operatorname{row}(Z_{s,m}^{t-1})}$ denotes the orthogonal projection
onto the row space of $Z_{s,m}^{t-1}$ and $\sigma_{\min}^{+}(Z_{s,m}^{t-1})$ its
smallest nonzero singular value.
\end{theorem}

\textbf{Jacobian.}
For fixed $(m,n,t)$, the Jacobian of a GAT can be viewed as a mapping
\[
\mathcal J_{mn}:(\alpha_m,Z)\mapsto J_{mn}(\alpha_m,Z),
\]
with the server and oracle Jacobians given by
\[
J^{(s)}_{mn}(t)
= \mathcal J_{mn}(\alpha_m^s, Z_{s,m}^{t-1}),
\qquad
J^{(o)}_{mn}(t)
= \mathcal J_{mn}(\alpha_m^o, Z_{o,m}^{t-1}).
\]
Explicitly, these Jacobians are defined as
\[
J^{(s)}_{mn}(t)
:=
\frac{\partial (\tilde{h}_s^t)_m}{\partial (\hat{h}_c^{t-1})_n},
\qquad
J^{(o)}_{mn}(t)
:=
\frac{\partial (\tilde{h}_o^t)_m}{\partial (\hat{h}_o^{t-1})_n}.
\]

\begin{assumption}[\textbf{Lipschitz continuity}]
There exists a constant $L_J>0$ such that for all $(\alpha,Z)$ and
$(\tilde \alpha,\tilde Z)$,
\[
\big\|
\mathcal J_{mn}(\alpha,Z)
-
\mathcal J_{mn}(\tilde \alpha,\tilde Z)
\big\|
\le
L_J\big(\|\alpha-\tilde \alpha\| + \|Z-\tilde Z\|_F\big).
\]
\end{assumption}

\begin{theorem}[\textbf{Bounded Jacobian}]
\label{thm:J_conv}
For each fixed $m$ and a neighboring client $n$, the difference between the Jacobian of
the oracle and server GAT models is bounded as, 
\begin{multline*}
\big\|J^{(s)}_{mn}(t) - J^{(o)}_{mn}(t)\big\|
\;\le\;
L_J\Bigg(
\frac{C_h\,\varepsilon_2 + L_{\sigma^{-1}}\,\varepsilon_1}
     {\sigma_{\min}^{+}(Z_{s,m}^{t-1})}
\\
+ \big\|P_{\ker(Z_{s,m}^{t-1})}(\alpha_m^o-\alpha_m^s)\big\|
+ \sqrt{N_m}\,\varepsilon_2
\Bigg).
\end{multline*}
\end{theorem}
The middle term lies in the null space of $Z_{s,m}^{t-1}$ and does not affect
the one-step predictions; it vanishes when $Z_{s,m}^{t-1}$ has full column rank.

\section{Experiments}\label{sec:Experiments}
\subsection{Setup}
\paragraph{\textbf{Datasets.}}
We evaluate our framework on both synthetic and real-world multivariate
time-series datasets. The synthetic experiments use three systems with
known temporal interdependencies: \textbf{(1)} a Linear SSM, \textbf{(2)} a Nonlinear SSM, and \textbf{(3)} the
Lorenz--96 chaotic dynamical system. All of these datasets use $M = 6$ clients, state dimension $p_m = 2 \hspace{0.1cm} \forall m$, and observation dimension $d_m = 8 \hspace{0.1cm} \forall m$. These datasets enable direct
comparison against the ground-truth temporal interdependencies and are
used to validate the theoretical claims in Sections~\ref{sec:Methodology}, \ref{sec:InterpretingJacobian}, and \ref{sec:Convergence}. We
further evaluate our method on three real-world benchmarks: the
\textbf{(1)} HAI industrial control system dataset \cite{github_HAI}, \textbf{(2)} the Tennessee Eastman Process
(TEP) \cite{TEP}, and \textbf{(3)} the Server Machine Dataset (SMD) \cite{SMD}, where the objective is
root-cause analysis under unknown interdependencies.

{\textbf{\textit{Baselines}.}}
We compare against recent federated causal discovery and graph learning
methods, including FedPC\cite{FedPC}, FedCSL\cite{FedCSL}, NOTEARS-ADMM\cite{ng2022fedbn}, FDBNL\cite{FDBNL}, and FedDAG\cite{gao2023feddag}. We
also compare against two reference models: \textbf{(i)} the Proprietary model,
which uses the client models without augmentation, and \textbf{(ii)} the
Centralized Oracle, which has access to observations from all clients.

{\textbf{\textit{Evaluation Metrics}.}}
Prediction accuracy is measured using the client test \textit{mean squared error}
(MSE) for both synthetic and real-world datasets. On the synthetic datasets, temporal interdependency estimation is
evaluated using the \textit{cosine similarity} between the learned and
ground-truth Jacobian operators. We additionally report the \textit{temporal correlation} between the learned
attention coefficients and Jacobian magnitudes. On the real-world datasets, the ground-truth temporal interdependencies are unavailable. We therefore evaluate the performance through the downstream task of root-cause analysis using AC@$k$ \cite{AERCA}, defined as the fraction of anomalies for which at least one ground-truth root-cause client is ranked within the top-$k$ predictions.


\subsection{Synthetic Validation}
\label{sec:Synthetic}
\begin{figure}[htbp]
    \centering
    \includegraphics[width=0.48\textwidth]{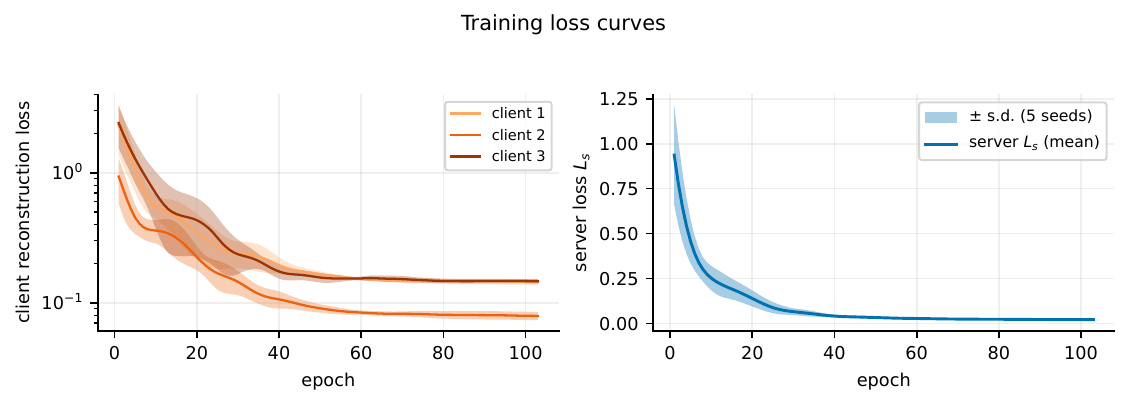}
    \caption{Training loss curves on the nonlinear SSM synthetic dataset for clients (left) and server (right) for a M=3 clients.}
    \label{fig:training}
\end{figure}

\paragraph{\textbf{Training Dynamics}.}
Figure~\ref{fig:training} illustrates the client and server training
losses on the nonlinear SSM dataset for a toy example with $M=3$ clients (Appendix \ref{app:synth-m3}). Both objectives
converge smoothly, demonstrating stable joint optimization of the local
client models and the server transition model. 

\textbf{\textit{Prediction Accuracy.}}
Table~\ref{tab:prediction} compares the proprietary client models, ourframework, and the centralized oracle. Consistent with
Claims~\ref{claim1}--\ref{claim2}, the proposed framework consistently
closes a large portion of the gap between the proprietary models and the
centralized oracle across all three synthetic systems, demonstrating the
benefit of learning interdependencies through augmentation. 

\begin{figure}[htbp]
    \centering
    \includegraphics[width=0.48\textwidth]{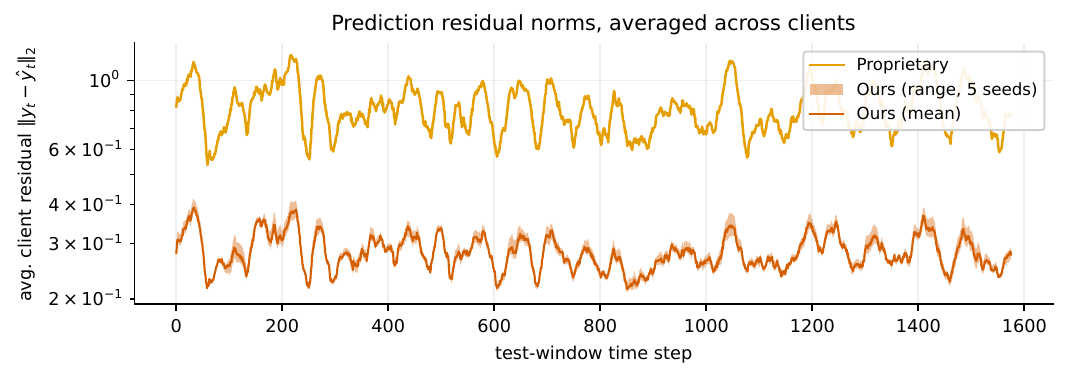}
    \caption{Average client residuals on the nonlinear SSM.}
    \label{fig:client_residuals}
\end{figure}

\begin{table}[H]
\centering
\caption{Prediction accuracy (test MSE) on synthetic datasets.}
\label{tab:prediction}
\setlength{\tabcolsep}{2pt}
\renewcommand{\arraystretch}{0.75}
\begin{tabular}{lccc}
\toprule
\textbf{Method} &
\textbf{Linear SSM} &
\textbf{Nonlinear SSM} &
\textbf{Lorenz-96} \\
\midrule
Centralized Oracle & 0.036 & 0.005 & 0.001 \\
Proprietary & 0.654 & 0.117 & 0.932 \\
\textbf{Ours} & \textbf{0.101} & \textbf{0.010} & \textbf{0.011} \\
\midrule
Reduction (\%) & 84.6 & 91.4 & 98.9 \\
\bottomrule
\end{tabular}
\end{table}
For the same 3-client toy example as Appendix~\ref{app:synth-m3}, on the nonlinear SSM dataset, Figure~\ref{fig:client_residuals} further shows that our model follows the proprietary model's temporal residual pattern but with consistently lower magnitude.

\textbf{\textit{Theoretical Validation}.}
Table~\ref{tab:structure} compares the learned temporal
interdependencies against the ground-truth Jacobian using
cosine similarity. Our framework consistently achieves the
highest agreement across all three datasets, providing empirical support
for theoretical results in Section \ref{sec:Convergence}. 
\begin{table}[htbp]
\centering
\caption{Cosine similarity: learned vs ground truth Jacobians.}
\label{tab:structure}
\setlength{\tabcolsep}{2pt}
\renewcommand{\arraystretch}{0.75}
{\small
\begin{tabular}{lccc}
\toprule
\textbf{Method} &
\textbf{Linear SSM} &
\textbf{Nonlinear SSM} &
\textbf{Lorenz-96} \\
\midrule
FedPC & 0.072 & 0.586 & 0.499 \\
FedCSL & 0.072 & 0.585 & 0.499 \\
NOTEARS-ADMM & 0.068 & 0.583 & 0.497 \\
FDBNL & 0.063 & 0.587 & 0.500 \\
FedDAG & 0.027 & 0.415 & 0.477 \\
\textbf{Ours} & \textbf{0.965} & \textbf{0.956} & \textbf{0.732} \\
\bottomrule
\end{tabular}}
\end{table}

\begin{figure}[htbp]
\centering
\includegraphics[width=\linewidth]{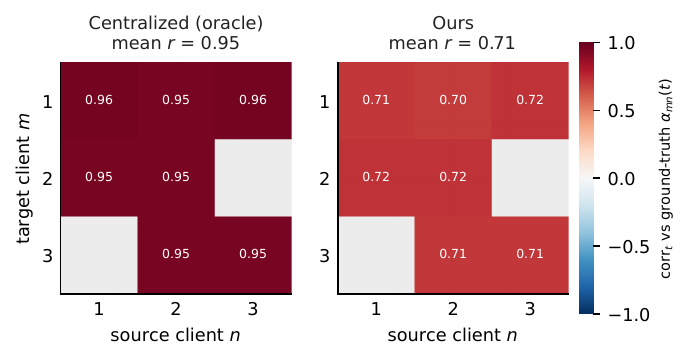}
{(a) Attention coefficients.}
\includegraphics[width=\linewidth]{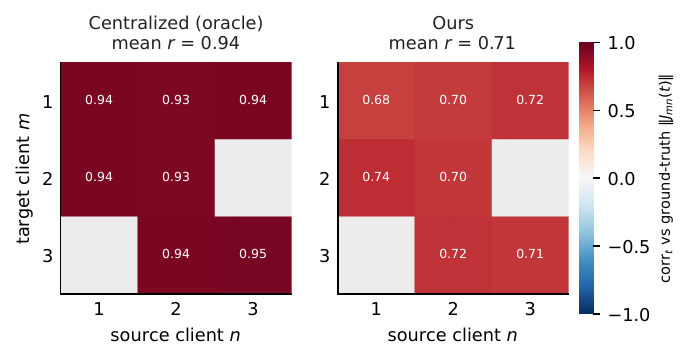}
{(b) Jacobian operators.}
\includegraphics[width=\linewidth]{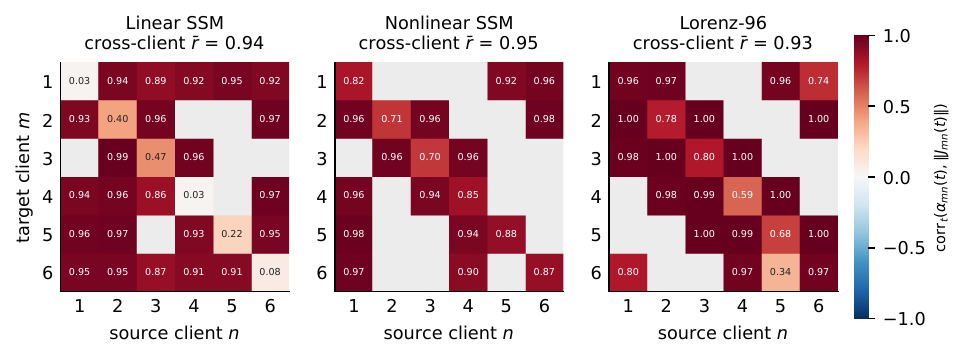}
{(c) Attention--Jacobian correlation.}

\caption{Synthetic validation: Correlation between the learned model and the ground truth for (a) attention coefficients and (b) Jacobian operators. (c) Correlation between attention coefficients and Jacobians in the learned model.}
\label{fig:interpretability}
\end{figure}

Furthermore, Figure~\ref{fig:interpretability} provides a detailed empirical
validation of the theoretical analysis in
Section~\ref{sec:InterpretingJacobian}. Figures~\ref{fig:interpretability}(a)--(b)
demonstrate agreement between the learned attention coefficients and
Jacobian operators and their centralized oracle counterparts, validating
Theorems~\ref{thm:param_conv} and
\ref{thm:J_conv}. Figure~\ref{fig:interpretability}(c)
empirically verifies the analytical relationship established in
Proposition~\ref{prop:jacobian_gat}. 

\subsection{Real-world Experiments}
\label{sec:RealWorld}
\textbf{\textit{Root Cause Analysis (RCA)}.}
Table~\ref{tab:rca} reports AC@$k$ on all three datasets. Across all
benchmarks, our proposed framework consistently outperforms the
proprietary models and competing federated graph learning baselines. Further explanation of the differences observed across all three datasets is provided in the Appendix. 

\begin{table}[htbp]
\centering
\caption{RCA performance (AC@$k$) on real-world datasets.}
\label{tab:rca}
\setlength{\tabcolsep}{2pt}
\renewcommand{\arraystretch}{1}
{\small
\begin{tabular}{lcc|cc|cc}
\hline
& \multicolumn{2}{c|}{\textbf{HAI}}
& \multicolumn{2}{c|}{\textbf{TEP}}
& \multicolumn{2}{c}{\textbf{SMD}} \\
\textbf{Method}
& \textbf{AC@1}
& \textbf{AC@2}
& \textbf{AC@1}
& \textbf{AC@2}
& \textbf{AC@1}
& \textbf{AC@2} \\
\hline
FedPC           & 0.41 & 0.76 & 0.19 & 0.40 & 0.41 & 0.60 \\
FedCSL          & 0.41 & 0.65 & 0.20 & 0.39 & 0.43 & 0.63 \\
NOTEARS-ADMM    & 0.35 & 0.76 & 0.18 & 0.35 & 0.42 & 0.61 \\
FDBNL           & 0.41 & 0.71 & 0.19 & 0.37 & 0.44 & 0.62 \\
FedDAG          & 0.35 & 0.65 & 0.18 & 0.39 & 0.41 & 0.62 \\
\textbf{Ours}   & \textbf{0.76} & \textbf{0.94}
                & \textbf{0.45} & \textbf{0.65}
                & \textbf{0.53} & \textbf{0.70} \\
\hline
\end{tabular}}
\end{table}
\textbf{\textit{Prediction Accuracy}.}
Table~\ref{tab:mse} reports the client prediction MSE on the three
real-world datasets. Consistent with the synthetic experiments, our
framework consistently improves over the proprietary client
models while approaching the centralized oracle. The largest gains are
observed on HAI and SMD datasets, where modeling cross-client temporal
interdependencies provides substantial predictive benefit. Please refer to the Appendix for details. 

\begin{table}[htbp]
\centering
\caption{Prediction accuracy (test MSE) on real-world datasets.}
\label{tab:mse}
\setlength{\tabcolsep}{6pt}
\renewcommand{\arraystretch}{1}
{\small
\begin{tabular}{lccc}
\hline
\textbf{Method} & \textbf{HAI} & \textbf{TEP} & \textbf{SMD} \\
\hline
Centralized (Oracle) & 0.240 & 0.635 & 0.377 \\
Proprietary          & 0.873 & 0.720 & 0.712 \\
\textbf{Ours}        & \textbf{0.508} & \textbf{0.699} & \textbf{0.558} \\
\hline
Reduction (\%) & 41.8 & 2.9 & 21.6 \\
\hline
\end{tabular}}
\end{table}

\subsection{Ablation Studies}\label{sec:Ablation}

\textbf{\textit{Robustness to Client Model Quality}.}
To evaluate robustness, at the clients, we progressively degrade the quality of the
proprietary models by replacing a subset of clients with randomly
initialized models. Table~\ref{tab:robustness} shows that the proposed
framework consistently outperforms the proprietary models and maintains
strong root-cause localization even when multiple client models are
corrupted, demonstrating that useful cross-client temporal
interdependencies can still be learned under severe local model
degradation.

\begin{table}[htbp]
\centering
\caption{Robustness under corrupted proprietary models.}
\label{tab:robustness}
\setlength{\tabcolsep}{4pt}
\renewcommand{\arraystretch}{1}
{\small
\begin{tabular}{llcc}
\hline
\textbf{Dataset} & \textbf{Corrupted Clients} & \textbf{AC@1} & \textbf{AC@2} \\
\hline
\multirow{4}{*}{HAI}
& None               & 0.76 & 0.94 \\
& 1 of 3 (33\%)      & 0.59 & 0.94 \\
& 2 of 3 (67\%)      & 0.59 & 0.94 \\
& All       & 0.47 & 0.82 \\
\hline
\multirow{5}{*}{TEP}
& None               & 0.45 & 0.65 \\
& 1 of 5 (20\%)      & 0.40 & 0.63 \\
& 2 of 5 (40\%)      & 0.38 & 0.66 \\
& 3 of 5 (60\%)      & 0.39 & 0.54 \\
& All       & 0.27 & 0.53 \\
\hline
\multirow{5}{*}{SMD}
& None               & 0.53 & 0.70 \\
& 1 of 5 (20\%)      & 0.51 & 0.68 \\
& 2 of 5 (40\%)      & 0.53 & 0.67 \\
& 3 of 5 (60\%)      & 0.52 & 0.68 \\
& All       & 0.54 & 0.68 \\
\hline
\end{tabular}}
\end{table}

\textbf{\textit{Server Architecture}.}
To justify the graph-attention server, we replace it with graph-blind
alternatives while keeping the federated learning protocol unchanged.
We report the fraction of off-diagonal Jacobian mass assigned to absent
edges. A graph-blind uniform allocation would assign 0.233 on the
linear SSM, 0.567 on the nonlinear SSM, and 0.400 on the Lorenz--96
dataset. Because GAT masks absent edges by construction, this quantity
is identically zero.
Table~\ref{tab:server} shows that GAT achieves the best prediction
accuracy while assigning virtually no influence to nonexistent edges. This
demonstrates that incorporating graph structure substantially improves
both prediction and interpretability.

\begin{table}[H]
\centering
\caption{Comparison of server architectures.}
\label{tab:server}
\setlength{\tabcolsep}{3pt}
\renewcommand{\arraystretch}{1}
{\small
\begin{tabular}{llcc}
\hline
\textbf{Dataset} & \textbf{Server} & \textbf{Client Loss} $\downarrow$ & \textbf{Mass on Absent Edges} $\downarrow$ \\
\hline
\multirow{4}{*}{Linear SSM}
& \textbf{GAT} & \textbf{0.101} & \textbf{0.000} \\
& LSTM         & 0.101 & 0.146 \\
& MLP          & 0.101 & 0.129 \\
& Transformer  & 0.101 & 0.224 \\
\hline
\multirow{4}{*}{Nonlinear SSM}
& \textbf{GAT} & \textbf{0.0100} & \textbf{0.000} \\
& LSTM         & 0.0112 & 0.457 \\
& MLP          & 0.0131 & 0.369 \\
& Transformer  & 0.0136 & 0.552 \\
\hline
\multirow{4}{*}{Lorenz-96}
& \textbf{GAT} & \textbf{0.0106} & \textbf{0.000} \\
& LSTM         & 0.0444 & 0.281 \\
& MLP          & 0.1132 & 0.404 \\
& Transformer  & 0.1761 & 0.383 \\
\hline
\end{tabular}}
\end{table}

\textbf{\textit{Graph Misspecification}.}
The proposed framework assumes \textit{a priori} knowledge of the graph skeleton. To evaluate
its robustness to graph errors, we progressively perturb the server
graph while keeping the data-generating process unchanged.
Table~\ref{tab:adjacency} shows that prediction accuracy remains stable,
whereas interpretability degrades gracefully as graph misspecification
increases. A fully connected graph consistently outperforms an
over-pruned graph, suggesting that dense connectivity is a safer choice
when prior graph knowledge is unavailable.

\begin{table}[htbp]
\centering
\caption{Graph misspecification study on synthetic datasets.}
\label{tab:adjacency}
\setlength{\tabcolsep}{2pt}
\renewcommand{\arraystretch}{1}
{\small
\begin{tabular}{llcc}
\toprule
\textbf{Dataset} & \textbf{Server Graph} &
\textbf{Client Loss} $\downarrow$ &
\textbf{Cosine Similarity} $\uparrow$ \\
\midrule
\multirow{6}{*}{Linear SSM}
& True Graph      & 0.1010 & 0.965 \\
& $\rho=10\%$     & 0.1010 & $0.880 \pm 0.036$ \\
& $\rho=20\%$     & 0.1010 & $0.817 \pm 0.026$ \\
& $\rho=30\%$     & 0.1010 & $0.721 \pm 0.019$ \\
& Fully Connected & 0.1010 & 0.932 \\
& Diagonal Only   & 0.1000 & 0.067 \\
\midrule
\multirow{6}{*}{Nonlinear SSM}
& True Graph      & 0.01027 & 0.956 \\
& $\rho=10\%$     & 0.01016 & $0.925 \pm 0.008$ \\
& $\rho=20\%$     & 0.01018 & $0.846 \pm 0.018$ \\
& $\rho=30\%$     & 0.01020 & $0.828 \pm 0.008$ \\
& Fully Connected & 0.01014 & 0.890 \\
& Diagonal Only   & 0.01003 & 0.587 \\
\midrule
\multirow{6}{*}{Lorenz--96}
& True Graph      & 0.0109 & 0.732 \\
& $\rho=10\%$     & 0.0109 & $0.720 \pm 0.024$ \\
& $\rho=20\%$     & 0.0109 & $0.650 \pm 0.022$ \\
& $\rho=30\%$     & 0.0109 & $0.632 \pm 0.005$ \\
& Fully Connected & 0.0108 & 0.710 \\
& Diagonal Only   & 0.0107 & 0.498 \\
\bottomrule
\end{tabular}}
\end{table}



\subsection{Scalability Studies}\label{sec:scalability}

To evaluate scalability, we systematically vary three key problem dimensions while keeping all other settings fixed: (i) the observation dimension $d_m$, (ii) the latent state dimension $p_m$, and (iii) the number of clients $M$ in the decentralized system. For each setting, we measure the server loss to assess whether the proposed framework maintains stable performance as the problem size increases. Due to space constraints, the results are provided in Appendix \ref{appendix:scalability}.

\subsection{Privacy Analysis}\label{sec:privacy}

We provide a differential privacy analysis under the Gaussian mechanism. While a comprehensive treatment is beyond the scope of this paper, Appendix \ref{appendix:privacy} includes rigorous theoretical guarantees and Appendix \ref{appendix:evaluation} provides empirical privacy--utility evaluation.

\section{Limitations}\label{sec:Limitations}
The framework assumes a known graph skeleton, which may be unavailable in practice, and motivates integration with causal or structure discovery methods. Additionally, the Jacobian-based interpretability relies on local first-order approximations and may degrade under strong nonlinearities. Furthermore, the present implementation assumes synchronous communication among participating clients. Finally, the theoretical privacy guarantees hold under the stated sensitivity and clipping assumptions. The empirical evaluation does not establish worst-case sensitivity bounds. 

\bibliographystyle{ACM-Reference-Format}
\bibliography{sample-base}


\begin{thebibliography}{46}


\ifx \showCODEN    \undefined \def \showCODEN     #1{\unskip}     \fi
\ifx \showISBNx    \undefined \def \showISBNx     #1{\unskip}     \fi
\ifx \showISBNxiii \undefined \def \showISBNxiii  #1{\unskip}     \fi
\ifx \showISSN     \undefined \def \showISSN      #1{\unskip}     \fi
\ifx \showLCCN     \undefined \def \showLCCN      #1{\unskip}     \fi
\ifx \shownote     \undefined \def \shownote      #1{#1}          \fi
\ifx \showarticletitle \undefined \def \showarticletitle #1{#1}   \fi
\ifx \showURL      \undefined \def \showURL       {\relax}        \fi
\providecommand\bibfield[2]{#2}
\providecommand\bibinfo[2]{#2}
\providecommand\natexlab[1]{#1}
\providecommand\showeprint[2][]{arXiv:#2}

\bibitem[Abadi et~al\mbox{.}(2016)]%
        {DPSGD}
\bibfield{author}{\bibinfo{person}{Martin Abadi}, \bibinfo{person}{Andy Chu}, \bibinfo{person}{Ian Goodfellow}, \bibinfo{person}{H.~Brendan McMahan}, \bibinfo{person}{Ilya Mironov}, \bibinfo{person}{Kunal Talwar}, {and} \bibinfo{person}{Li Zhang}.} \bibinfo{year}{2016}\natexlab{}.
\newblock \showarticletitle{Deep Learning with Differential Privacy}. In \bibinfo{booktitle}{\emph{Proceedings of the 2016 ACM SIGSAC Conference on Computer and Communications Security}} (Vienna, Austria) \emph{(\bibinfo{series}{CCS '16})}. \bibinfo{publisher}{Association for Computing Machinery}, \bibinfo{address}{New York, NY, USA}, \bibinfo{pages}{308–318}.
\newblock
\showISBNx{9781450341394}


\bibitem[Battaglia et~al\mbox{.}(2016)]%
        {battaglia2016interaction}
\bibfield{author}{\bibinfo{person}{Peter~W. Battaglia}, \bibinfo{person}{Razvan Pascanu}, \bibinfo{person}{Matthew Lai}, \bibinfo{person}{Danilo~Jimenez Rezende}, {and} \bibinfo{person}{Koray Kavukcuoglu}.} \bibinfo{year}{2016}\natexlab{}.
\newblock \showarticletitle{Interaction Networks for Learning about Objects, Relations and Physics}. In \bibinfo{booktitle}{\emph{Advances in Neural Information Processing Systems}}.
\newblock


\bibitem[Brody et~al\mbox{.}(2022)]%
        {GATv2}
\bibfield{author}{\bibinfo{person}{Shaked Brody}, \bibinfo{person}{Uri Alon}, {and} \bibinfo{person}{Eran Yahav}.} \bibinfo{year}{2022}\natexlab{}.
\newblock \showarticletitle{How Attentive Are Graph Attention Networks?}. In \bibinfo{booktitle}{\emph{International Conference on Learning Representations}}.
\newblock


\bibitem[Chen et~al\mbox{.}(2022)]%
        {chen2022vfg}
\bibfield{author}{\bibinfo{person}{Chaochao Chen}, \bibinfo{person}{Jun Zhou}, \bibinfo{person}{Longfei Zheng}, \bibinfo{person}{Huiwen Wu}, \bibinfo{person}{Lingjuan Lyu}, \bibinfo{person}{Jia Wu}, \bibinfo{person}{Bingzhe Wu}, \bibinfo{person}{Ziqi Liu}, \bibinfo{person}{Li Wang}, {and} \bibinfo{person}{Xiaolin Zheng}.} \bibinfo{year}{2022}\natexlab{}.
\newblock \showarticletitle{Vertically Federated Graph Neural Network for Privacy-Preserving Node Classification}. In \bibinfo{booktitle}{\emph{Proceedings of the Thirty-First International Joint Conference on Artificial Intelligence}}.
\newblock


\bibitem[Chen et~al\mbox{.}(2025)]%
        {FDBNL}
\bibfield{author}{\bibinfo{person}{Jianhong Chen}, \bibinfo{person}{Ying Ma}, {and} \bibinfo{person}{Xubo Yue}.} \bibinfo{year}{2025}\natexlab{}.
\newblock \showarticletitle{Federated Learning of Dynamic Bayesian Network via Continuous Optimization from Time Series Data}.
\newblock \bibinfo{journal}{\emph{IEEE Transactions on Artificial Intelligence}} (\bibinfo{year}{2025}).
\newblock


\bibitem[Chen et~al\mbox{.}(2023)]%
        {Cascading1}
\bibfield{author}{\bibinfo{person}{Lei Chen}, \bibinfo{person}{Sergey Gorbachev}, \bibinfo{person}{Dong Yue}, \bibinfo{person}{Chunxia Dou}, \bibinfo{person}{Xiangpeng Xie}, \bibinfo{person}{Shengquan Li}, \bibinfo{person}{Nan Zhao}, {and} \bibinfo{person}{Tingjun Zhang}.} \bibinfo{year}{2023}\natexlab{}.
\newblock \showarticletitle{Impact of cascading failure on power distribution and data transmission in cyber-physical power systems}.
\newblock \bibinfo{journal}{\emph{IEEE Transactions on Network Science and Engineering}} \bibinfo{volume}{11}, \bibinfo{number}{2} (\bibinfo{year}{2023}), \bibinfo{pages}{1580--1590}.
\newblock


\bibitem[Downs and Vogel(1993)]%
        {TEP}
\bibfield{author}{\bibinfo{person}{James~J Downs} {and} \bibinfo{person}{Ernest~F Vogel}.} \bibinfo{year}{1993}\natexlab{}.
\newblock \showarticletitle{A plant-wide industrial process control problem}.
\newblock \bibinfo{journal}{\emph{Computers \& chemical engineering}} \bibinfo{volume}{17}, \bibinfo{number}{3} (\bibinfo{year}{1993}), \bibinfo{pages}{245--255}.
\newblock


\bibitem[Dwork and Roth(2014)]%
        {DP_book}
\bibfield{author}{\bibinfo{person}{Cynthia Dwork} {and} \bibinfo{person}{Aaron Roth}.} \bibinfo{year}{2014}\natexlab{}.
\newblock \showarticletitle{The Algorithmic Foundations of Differential Privacy}.
\newblock \bibinfo{journal}{\emph{Foundations and Trends® in Theoretical Computer Science}} \bibinfo{volume}{9}, \bibinfo{number}{3–4} (\bibinfo{year}{2014}), \bibinfo{pages}{211--407}.
\newblock
\showISSN{1551-305X}


\bibitem[Gao et~al\mbox{.}(2023)]%
        {gao2023feddag}
\bibfield{author}{\bibinfo{person}{Erdun Gao}, \bibinfo{person}{Junjia Chen}, \bibinfo{person}{Li Shen}, \bibinfo{person}{Tongliang Liu}, \bibinfo{person}{Mingming Gong}, {and} \bibinfo{person}{Howard Bondell}.} \bibinfo{year}{2023}\natexlab{}.
\newblock \showarticletitle{{FedDAG}: Federated DAG Structure Learning}.
\newblock \bibinfo{journal}{\emph{Transactions on Machine Learning Research}} (\bibinfo{year}{2023}).
\newblock


\bibitem[Graber and Schwing(2020)]%
        {graber2020dnri}
\bibfield{author}{\bibinfo{person}{Colin Graber} {and} \bibinfo{person}{Alexander~G. Schwing}.} \bibinfo{year}{2020}\natexlab{}.
\newblock \showarticletitle{Dynamic Neural Relational Inference}. In \bibinfo{booktitle}{\emph{Proceedings of the IEEE/CVF Conference on Computer Vision and Pattern Recognition}}.
\newblock


\bibitem[Guo et~al\mbox{.}(2024)]%
        {FedCSL}
\bibfield{author}{\bibinfo{person}{Xianjie Guo}, \bibinfo{person}{Kui Yu}, \bibinfo{person}{Lin Liu}, {and} \bibinfo{person}{Jiuyong Li}.} \bibinfo{year}{2024}\natexlab{}.
\newblock \showarticletitle{FedCSL: A scalable and accurate approach to federated causal structure learning}. In \bibinfo{booktitle}{\emph{Proceedings of the AAAI Conference on Artificial Intelligence}}, Vol.~\bibinfo{volume}{38}. \bibinfo{pages}{12235--12243}.
\newblock


\bibitem[Han et~al\mbox{.}(2025)]%
        {AERCA}
\bibfield{author}{\bibinfo{person}{Xiao Han}, \bibinfo{person}{Saima Absar}, \bibinfo{person}{Lu Zhang}, {and} \bibinfo{person}{Shuhan Yuan}.} \bibinfo{year}{2025}\natexlab{}.
\newblock \showarticletitle{Root cause analysis of anomalies in multivariate time series through granger causal discovery}. In \bibinfo{booktitle}{\emph{The Thirteenth International Conference on Learning Representations}}.
\newblock


\bibitem[He et~al\mbox{.}(2021)]%
        {he2021fedgraphnn}
\bibfield{author}{\bibinfo{person}{Chaoyang He}, \bibinfo{person}{Keshav Balasubramanian}, \bibinfo{person}{Emir Ceyani}, \bibinfo{person}{Carl Yang}, \bibinfo{person}{Han Xie}, \bibinfo{person}{Lichao Sun}, \bibinfo{person}{Lifang He}, \bibinfo{person}{Liangwei Yang}, \bibinfo{person}{Philip~S. Yu}, \bibinfo{person}{Yu Rong}, \bibinfo{person}{Peilin Zhao}, \bibinfo{person}{Junzhou Huang}, \bibinfo{person}{Murali Annavaram}, {and} \bibinfo{person}{Salman Avestimehr}.} \bibinfo{year}{2021}\natexlab{}.
\newblock \bibinfo{title}{FedGraphNN: A Federated Learning System and Benchmark for Graph Neural Networks}.
\newblock
\showeprint[arxiv]{2104.07145}~[cs.LG]
\urldef\tempurl%
\url{https://arxiv.org/abs/2104.07145}
\showURL{%
\tempurl}


\bibitem[Huang et~al\mbox{.}(2023)]%
        {FedPC}
\bibfield{author}{\bibinfo{person}{Jianli Huang}, \bibinfo{person}{Xianjie Guo}, \bibinfo{person}{Kui Yu}, \bibinfo{person}{Fuyuan Cao}, {and} \bibinfo{person}{Jiye Liang}.} \bibinfo{year}{2023}\natexlab{}.
\newblock \showarticletitle{Towards privacy-aware causal structure learning in federated setting}.
\newblock \bibinfo{journal}{\emph{IEEE Transactions on Big Data}} \bibinfo{volume}{9}, \bibinfo{number}{6} (\bibinfo{year}{2023}), \bibinfo{pages}{1525--1535}.
\newblock


\bibitem[Kairouz et~al\mbox{.}(2021)]%
        {FL_review2}
\bibfield{author}{\bibinfo{person}{Peter Kairouz}, \bibinfo{person}{H~Brendan McMahan}, \bibinfo{person}{Brendan Avent}, \bibinfo{person}{Aur{\'e}lien Bellet}, \bibinfo{person}{Mehdi Bennis}, \bibinfo{person}{Arjun~Nitin Bhagoji}, \bibinfo{person}{Kallista Bonawitz}, \bibinfo{person}{Zachary Charles}, \bibinfo{person}{Graham Cormode}, \bibinfo{person}{Rachel Cummings}, {et~al\mbox{.}}} \bibinfo{year}{2021}\natexlab{}.
\newblock \showarticletitle{Advances and open problems in federated learning}.
\newblock \bibinfo{journal}{\emph{Foundations and trends{\textregistered} in machine learning}} \bibinfo{volume}{14}, \bibinfo{number}{1--2} (\bibinfo{year}{2021}), \bibinfo{pages}{1--210}.
\newblock


\bibitem[Kalisch(2007)]%
        {kalisch2007estimating}
\bibfield{author}{\bibinfo{person}{Markus Kalisch}.} \bibinfo{year}{2007}\natexlab{}.
\newblock \showarticletitle{Estimating High-Dimensional Directed Acyclic Graphs with the PC-Algorithm}.
\newblock \bibinfo{journal}{\emph{Journal of Machine Learning Research}}  \bibinfo{volume}{8} (\bibinfo{year}{2007}), \bibinfo{pages}{613--636}.
\newblock


\bibitem[Kipf et~al\mbox{.}(2018)]%
        {kipf2018nri}
\bibfield{author}{\bibinfo{person}{Thomas~N. Kipf}, \bibinfo{person}{Ethan Fetaya}, \bibinfo{person}{Kuan-Chieh Wang}, \bibinfo{person}{Max Welling}, {and} \bibinfo{person}{Richard Zemel}.} \bibinfo{year}{2018}\natexlab{}.
\newblock \showarticletitle{Neural Relational Inference for Interacting Systems}. In \bibinfo{booktitle}{\emph{Proceedings of the 35th International Conference on Machine Learning}}.
\newblock


\bibitem[Lachapelle et~al\mbox{.}(2020)]%
        {lachapelle2020grandag}
\bibfield{author}{\bibinfo{person}{S{\'e}bastien Lachapelle}, \bibinfo{person}{Philippe Brouillard}, \bibinfo{person}{Tristan Deleu}, {and} \bibinfo{person}{Simon Lacoste-Julien}.} \bibinfo{year}{2020}\natexlab{}.
\newblock \showarticletitle{Gradient-Based Neural DAG Learning}. In \bibinfo{booktitle}{\emph{Proceedings of the Eighth International Conference on Learning Representations}}.
\newblock


\bibitem[Li et~al\mbox{.}(2020)]%
        {FL_review1}
\bibfield{author}{\bibinfo{person}{Tian Li}, \bibinfo{person}{Anit~Kumar Sahu}, \bibinfo{person}{Ameet Talwalkar}, {and} \bibinfo{person}{Virginia Smith}.} \bibinfo{year}{2020}\natexlab{}.
\newblock \showarticletitle{Federated learning: Challenges, methods, and future directions}.
\newblock \bibinfo{journal}{\emph{IEEE signal processing magazine}} \bibinfo{volume}{37}, \bibinfo{number}{3} (\bibinfo{year}{2020}), \bibinfo{pages}{50--60}.
\newblock


\bibitem[Li et~al\mbox{.}(2018)]%
        {li2018dcrnn}
\bibfield{author}{\bibinfo{person}{Yaguang Li}, \bibinfo{person}{Rose Yu}, \bibinfo{person}{Cyrus Shahabi}, {and} \bibinfo{person}{Yan Liu}.} \bibinfo{year}{2018}\natexlab{}.
\newblock \showarticletitle{Diffusion Convolutional Recurrent Neural Network: Data-Driven Traffic Forecasting}. In \bibinfo{booktitle}{\emph{International Conference on Learning Representations}}.
\newblock


\bibitem[Ma et~al\mbox{.}(2023)]%
        {ma2023federated}
\bibfield{author}{\bibinfo{person}{Tengfei Ma}, \bibinfo{person}{Trong~Nghia Hoang}, {and} \bibinfo{person}{Jie Chen}.} \bibinfo{year}{2023}\natexlab{}.
\newblock \showarticletitle{Federated learning of models pre-trained on different features with consensus graphs}. In \bibinfo{booktitle}{\emph{Uncertainty in artificial intelligence}}. PMLR, \bibinfo{pages}{1336--1346}.
\newblock


\bibitem[Mai and Pang(2023)]%
        {mai2023verfedgnn}
\bibfield{author}{\bibinfo{person}{Peihua Mai} {and} \bibinfo{person}{Yan Pang}.} \bibinfo{year}{2023}\natexlab{}.
\newblock \showarticletitle{{VerFedGNN}: Vertical Federated Graph Neural Network for Recommender Systems}. In \bibinfo{booktitle}{\emph{Proceedings of the 40th International Conference on Machine Learning}}.
\newblock


\bibitem[Mironov(2017)]%
        {RDP}
\bibfield{author}{\bibinfo{person}{Ilya Mironov}.} \bibinfo{year}{2017}\natexlab{}.
\newblock \showarticletitle{Rényi Differential Privacy}. In \bibinfo{booktitle}{\emph{2017 IEEE 30th Computer Security Foundations Symposium (CSF)}}. \bibinfo{pages}{263--275}.
\newblock
\href{https://doi.org/10.1109/CSF.2017.11}{doi:\nolinkurl{10.1109/CSF.2017.11}}


\bibitem[Ng and Zhang(2022)]%
        {ng2022fedbn}
\bibfield{author}{\bibinfo{person}{Ignavier Ng} {and} \bibinfo{person}{Kun Zhang}.} \bibinfo{year}{2022}\natexlab{}.
\newblock \showarticletitle{Towards federated Bayesian network structure learning with continuous optimization}. In \bibinfo{booktitle}{\emph{International Conference on Artificial Intelligence and Statistics}}. PMLR, \bibinfo{pages}{8095--8111}.
\newblock


\bibitem[Pamfil et~al\mbox{.}(2020)]%
        {pamfil2020dynotears}
\bibfield{author}{\bibinfo{person}{Roxana Pamfil}, \bibinfo{person}{Nisara Sriwattanaworachai}, \bibinfo{person}{Shaan Desai}, \bibinfo{person}{Philip Pilgerstorfer}, \bibinfo{person}{Konstantinos Georgatzis}, \bibinfo{person}{Paul Beaumont}, {and} \bibinfo{person}{Bryon Aragam}.} \bibinfo{year}{2020}\natexlab{}.
\newblock \showarticletitle{DYNOTEARS: Structure Learning from Time-Series Data}. In \bibinfo{booktitle}{\emph{Proceedings of the 23rd International Conference on Artificial Intelligence and Statistics}}.
\newblock


\bibitem[Runge(2020)]%
        {runge2020discovering}
\bibfield{author}{\bibinfo{person}{Jakob Runge}.} \bibinfo{year}{2020}\natexlab{}.
\newblock \showarticletitle{Discovering contemporaneous and lagged causal relations in autocorrelated nonlinear time series datasets}. In \bibinfo{booktitle}{\emph{Conference on uncertainty in artificial intelligence}}. Pmlr, \bibinfo{pages}{1388--1397}.
\newblock


\bibitem[Runge et~al\mbox{.}(2019)]%
        {runge2019detecting}
\bibfield{author}{\bibinfo{person}{Jakob Runge}, \bibinfo{person}{Peer Nowack}, \bibinfo{person}{Marlene Kretschmer}, \bibinfo{person}{Seth Flaxman}, {and} \bibinfo{person}{Dino Sejdinovic}.} \bibinfo{year}{2019}\natexlab{}.
\newblock \showarticletitle{Detecting and quantifying causal associations in large nonlinear time series datasets}.
\newblock \bibinfo{journal}{\emph{Science advances}} \bibinfo{volume}{5}, \bibinfo{number}{11} (\bibinfo{year}{2019}), \bibinfo{pages}{eaau4996}.
\newblock


\bibitem[Shang et~al\mbox{.}(2021)]%
        {shang2021gts}
\bibfield{author}{\bibinfo{person}{Chao Shang}, \bibinfo{person}{Jie Chen}, {and} \bibinfo{person}{Jinbo Bi}.} \bibinfo{year}{2021}\natexlab{}.
\newblock \showarticletitle{Discrete Graph Structure Learning for Forecasting Multiple Time Series}. In \bibinfo{booktitle}{\emph{International Conference on Learning Representations}}.
\newblock


\bibitem[Shin et~al\mbox{.}(2014)]%
        {Cascading2}
\bibfield{author}{\bibinfo{person}{Dong-Hoon Shin}, \bibinfo{person}{Dajun Qian}, {and} \bibinfo{person}{Junshan Zhang}.} \bibinfo{year}{2014}\natexlab{}.
\newblock \showarticletitle{Cascading effects in interdependent networks}.
\newblock \bibinfo{journal}{\emph{Ieee Network}} \bibinfo{volume}{28}, \bibinfo{number}{4} (\bibinfo{year}{2014}), \bibinfo{pages}{82--87}.
\newblock


\bibitem[Shin et~al\mbox{.}(2023)]%
        {github_HAI}
\bibfield{author}{\bibinfo{person}{Hyeok-Ki Shin}, \bibinfo{person}{Woomyo Lee}, \bibinfo{person}{Seungoh Choi}, \bibinfo{person}{Jeong-Han Yun}, {and} \bibinfo{person}{Byung-Gi Min}.} \bibinfo{year}{2023}\natexlab{}.
\newblock \bibinfo{title}{HAI security datasets}.
\newblock
\urldef\tempurl%
\url{https://github.com/icsdataset/hai}
\showURL{%
\tempurl}


\bibitem[Stouffer et~al\mbox{.}(2011)]%
        {NIST_ICS}
\bibfield{author}{\bibinfo{person}{Keith Stouffer}, \bibinfo{person}{Joe Falco}, \bibinfo{person}{Karen Scarfone}, {et~al\mbox{.}}} \bibinfo{year}{2011}\natexlab{}.
\newblock \showarticletitle{Guide to industrial control systems (ICS) security}.
\newblock \bibinfo{journal}{\emph{NIST special publication}} \bibinfo{volume}{800}, \bibinfo{number}{82} (\bibinfo{year}{2011}), \bibinfo{pages}{16--16}.
\newblock


\bibitem[Stouffer et~al\mbox{.}(2023)]%
        {NIST_OT}
\bibfield{author}{\bibinfo{person}{Keith Stouffer}, \bibinfo{person}{Keith Stouffer}, \bibinfo{person}{Michael Pease}, \bibinfo{person}{CheeYee Tang}, \bibinfo{person}{Timothy Zimmerman}, \bibinfo{person}{Victoria Pillitteri}, \bibinfo{person}{Suzanne Lightman}, \bibinfo{person}{Adam Hahn}, \bibinfo{person}{Stephanie Saravia}, \bibinfo{person}{Aslam Sherule}, {et~al\mbox{.}}} \bibinfo{year}{2023}\natexlab{}.
\newblock \showarticletitle{Guide to operational technology (ot) security}.
\newblock  (\bibinfo{year}{2023}).
\newblock


\bibitem[Su et~al\mbox{.}(2019)]%
        {SMD}
\bibfield{author}{\bibinfo{person}{Ya Su}, \bibinfo{person}{Youjian Zhao}, \bibinfo{person}{Chenhao Niu}, \bibinfo{person}{Rong Liu}, \bibinfo{person}{Wei Sun}, {and} \bibinfo{person}{Dan Pei}.} \bibinfo{year}{2019}\natexlab{}.
\newblock \showarticletitle{Robust anomaly detection for multivariate time series through stochastic recurrent neural network}. In \bibinfo{booktitle}{\emph{Proceedings of the 25th ACM SIGKDD international conference on knowledge discovery \& data mining}}. \bibinfo{pages}{2828--2837}.
\newblock


\bibitem[Tan et~al\mbox{.}(2025)]%
        {DistDataCenters}
\bibfield{author}{\bibinfo{person}{Haobin Tan}, \bibinfo{person}{Yao Xiao}, \bibinfo{person}{Amelie~Chi Zhou}, \bibinfo{person}{Kezhong Lu}, {and} \bibinfo{person}{Xuan Yang}.} \bibinfo{year}{2025}\natexlab{}.
\newblock \showarticletitle{Distributed and Adaptive Partitioning for Large Graphs in Geo-Distributed Data Centers}.
\newblock \bibinfo{journal}{\emph{IEEE Transactions on Parallel and Distributed Systems}} (\bibinfo{year}{2025}).
\newblock


\bibitem[Veli{\v{c}}kovi{\'c} et~al\mbox{.}(2018a)]%
        {GATv1}
\bibfield{author}{\bibinfo{person}{Petar Veli{\v{c}}kovi{\'c}}, \bibinfo{person}{Guillem Cucurull}, \bibinfo{person}{Arantxa Casanova}, \bibinfo{person}{Adriana Romero}, \bibinfo{person}{Pietro Li{\`o}}, {and} \bibinfo{person}{Yoshua Bengio}.} \bibinfo{year}{2018}\natexlab{a}.
\newblock \showarticletitle{Graph Attention Networks}. In \bibinfo{booktitle}{\emph{International Conference on Learning Representations}}.
\newblock


\bibitem[Veli{\v{c}}kovi{\'c} et~al\mbox{.}(2018b)]%
        {velickovic2018gat}
\bibfield{author}{\bibinfo{person}{Petar Veli{\v{c}}kovi{\'c}}, \bibinfo{person}{Guillem Cucurull}, \bibinfo{person}{Arantxa Casanova}, \bibinfo{person}{Adriana Romero}, \bibinfo{person}{Pietro Li{\`o}}, {and} \bibinfo{person}{Yoshua Bengio}.} \bibinfo{year}{2018}\natexlab{b}.
\newblock \showarticletitle{Graph Attention Networks}. In \bibinfo{booktitle}{\emph{International Conference on Learning Representations}}.
\newblock


\bibitem[Wang et~al\mbox{.}(2024)]%
        {DistManuf}
\bibfield{author}{\bibinfo{person}{Hao Wang}, \bibinfo{person}{Hao Luo}, \bibinfo{person}{Lei Ren}, \bibinfo{person}{Mingyi Huo}, \bibinfo{person}{Yuchen Jiang}, {and} \bibinfo{person}{Okyay Kaynak}.} \bibinfo{year}{2024}\natexlab{}.
\newblock \showarticletitle{Data-driven design of distributed monitoring and optimization system for manufacturing systems}.
\newblock \bibinfo{journal}{\emph{IEEE Transactions on Industrial Informatics}} \bibinfo{volume}{20}, \bibinfo{number}{7} (\bibinfo{year}{2024}), \bibinfo{pages}{9455--9464}.
\newblock


\bibitem[Wu et~al\mbox{.}(2022)]%
        {wu2022fedpergnn}
\bibfield{author}{\bibinfo{person}{Chuhan Wu}, \bibinfo{person}{Fangzhao Wu}, \bibinfo{person}{Lingjuan Lyu}, \bibinfo{person}{Tao Qi}, \bibinfo{person}{Yongfeng Huang}, {and} \bibinfo{person}{Xing Xie}.} \bibinfo{year}{2022}\natexlab{}.
\newblock \showarticletitle{A federated graph neural network framework for privacy-preserving personalization}.
\newblock \bibinfo{journal}{\emph{Nature Communications}} \bibinfo{volume}{13}, \bibinfo{number}{1} (\bibinfo{year}{2022}), \bibinfo{pages}{3091}.
\newblock


\bibitem[Wu et~al\mbox{.}(2020)]%
        {wu2020mtgnn}
\bibfield{author}{\bibinfo{person}{Zonghan Wu}, \bibinfo{person}{Shirui Pan}, \bibinfo{person}{Guodong Long}, \bibinfo{person}{Jing Jiang}, \bibinfo{person}{Xiaojun Chang}, {and} \bibinfo{person}{Chengqi Zhang}.} \bibinfo{year}{2020}\natexlab{}.
\newblock \showarticletitle{Connecting the Dots: Multivariate Time Series Forecasting with Graph Neural Networks}. In \bibinfo{booktitle}{\emph{Proceedings of the 26th ACM SIGKDD International Conference on Knowledge Discovery and Data Mining}}.
\newblock


\bibitem[Wu et~al\mbox{.}(2019)]%
        {wu2019graphwavenet}
\bibfield{author}{\bibinfo{person}{Zonghan Wu}, \bibinfo{person}{Shirui Pan}, \bibinfo{person}{Guodong Long}, \bibinfo{person}{Jing Jiang}, {and} \bibinfo{person}{Chengqi Zhang}.} \bibinfo{year}{2019}\natexlab{}.
\newblock \showarticletitle{Graph WaveNet for Deep Spatial-Temporal Graph Modeling}. In \bibinfo{booktitle}{\emph{Proceedings of the Twenty-Eighth International Joint Conference on Artificial Intelligence}}.
\newblock


\bibitem[Xiong et~al\mbox{.}(2025)]%
        {DRLgrid}
\bibfield{author}{\bibinfo{person}{Luolin Xiong}, \bibinfo{person}{Anshul Goyal}, \bibinfo{person}{Kankar Bhattacharya}, \bibinfo{person}{Yang Tang}, \bibinfo{person}{Zhaoyang Dong}, \bibinfo{person}{Feng Qian}, {and} \bibinfo{person}{Venkata~Balaji Thummalacherla}.} \bibinfo{year}{2025}\natexlab{}.
\newblock \showarticletitle{DRL-Based Distributed Coordination of ISO and DSOs in Bi-Level Electricity Markets}.
\newblock \bibinfo{journal}{\emph{IEEE Transactions on Industrial Informatics}} (\bibinfo{year}{2025}).
\newblock


\bibitem[Yu et~al\mbox{.}(2018)]%
        {yu2018stgcn}
\bibfield{author}{\bibinfo{person}{Bing Yu}, \bibinfo{person}{Haoteng Yin}, {and} \bibinfo{person}{Zhanxing Zhu}.} \bibinfo{year}{2018}\natexlab{}.
\newblock \showarticletitle{Spatio-Temporal Graph Convolutional Networks: A Deep Learning Framework for Traffic Forecasting}. In \bibinfo{booktitle}{\emph{Proceedings of the Twenty-Seventh International Joint Conference on Artificial Intelligence}}.
\newblock


\bibitem[Yu et~al\mbox{.}(2019)]%
        {yu2019daggnn}
\bibfield{author}{\bibinfo{person}{Yue Yu}, \bibinfo{person}{Jie Chen}, \bibinfo{person}{Tian Gao}, {and} \bibinfo{person}{Mo Yu}.} \bibinfo{year}{2019}\natexlab{}.
\newblock \showarticletitle{DAG-GNN: DAG Structure Learning with Graph Neural Networks}. In \bibinfo{booktitle}{\emph{Proceedings of the 36th International Conference on Machine Learning}}.
\newblock


\bibitem[Zhang et~al\mbox{.}(2021)]%
        {zhang2021subgraphfl}
\bibfield{author}{\bibinfo{person}{Ke Zhang}, \bibinfo{person}{Carl Yang}, \bibinfo{person}{Xiaoxiao Li}, \bibinfo{person}{Lichao Sun}, {and} \bibinfo{person}{Siu~Ming Yiu}.} \bibinfo{year}{2021}\natexlab{}.
\newblock \showarticletitle{Subgraph federated learning with missing neighbor generation}.
\newblock \bibinfo{journal}{\emph{Advances in neural information processing systems}}  \bibinfo{volume}{34} (\bibinfo{year}{2021}), \bibinfo{pages}{6671--6682}.
\newblock


\bibitem[Zheng et~al\mbox{.}(2021)]%
        {zheng2021asfgnn}
\bibfield{author}{\bibinfo{person}{Longfei Zheng}, \bibinfo{person}{Jun Zhou}, \bibinfo{person}{Chaochao Chen}, \bibinfo{person}{Bingzhe Wu}, \bibinfo{person}{Li Wang}, {and} \bibinfo{person}{Benyu Zhang}.} \bibinfo{year}{2021}\natexlab{}.
\newblock \showarticletitle{ASFGNN: Automated Separated-Federated Graph Neural Network}.
\newblock \bibinfo{journal}{\emph{Peer-to-Peer Networking and Applications}}  \bibinfo{volume}{14} (\bibinfo{year}{2021}), \bibinfo{pages}{1692--1704}.
\newblock


\bibitem[Zheng et~al\mbox{.}(2018)]%
        {zheng2018notears}
\bibfield{author}{\bibinfo{person}{Xun Zheng}, \bibinfo{person}{Bryon Aragam}, \bibinfo{person}{Pradeep~K. Ravikumar}, {and} \bibinfo{person}{Eric~P. Xing}.} \bibinfo{year}{2018}\natexlab{}.
\newblock \showarticletitle{DAGs with {NO TEARS}: Continuous Optimization for Structure Learning}. In \bibinfo{booktitle}{\emph{Advances in Neural Information Processing Systems}}.
\newblock


\end{thebibliography}

\begin{center}
    \huge\bfseries Appendix
\end{center}
\vspace{0.5em}
\appendix

\providecommand{\rev}[1]{\textcolor{red}{#1}}
\section{Synthetic Dataset Details}
\label{app:synthetic}
All three synthetic benchmarks share the same interface: $M=6$ clients, a
latent state of dimension $p_m=2$ per client (joint state dimension $12$),
and $d_m=8$ observations per client ($48$ in total), over $T=8000$ time
steps with an 80/20 train-test split. Observations are
generated by a fixed client-specific linear measurement model.
Each dataset is generated once from a fixed data seed; all reported model
results are means over five independent training seeds.

\paragraph{Linear SSM.}
The state evolves as
\[
x_{t+1} = A_{\mathrm{lin}}\, x_t + w_t,
\]
where $w_t \sim \mathcal N(0, \sigma_q^2 I)$ with $\sigma_q = 0.15$.
The cross-client skeleton is a random directed graph in which each
off-diagonal client pair is connected with probability $0.75$ (self-loops
always present), yielding a dense graph with $23$ of the $30$ possible
cross-client edges. $A_{\mathrm{lin}}$ is built from $2\times 2$
standard-normal blocks placed on this skeleton; the off-diagonal
(cross-client) blocks are amplified by a factor of $6$, and the full matrix
is then rescaled to spectral radius $0.9$. This construction places most
of the one-step dynamics on the cross-client blocks, so each client's next
state is driven primarily by its neighbors, information that no local
model can access from its own observation history. The
ground-truth interdependency operator for Table~\ref{tab:structure} is
$A_{\mathrm{lin}}$ itself (constant in time).

\paragraph{Nonlinear SSM.}
The state evolves as
\[
x_{t+1} = \tanh\!\big(\mathrm{GAT}(x_t;\, A)\big) + w_t,
\]
with $\sigma_q = 0.05$, where the state-transition function is itself a one-layer graph attention
network with predetermined parameters, serving as ground truth: a shared
$2\times2$ value transform, additive attention with a
$\mathrm{LeakyReLU}$ scoring function and a logit-sharpening factor,
and a masked softmax over the given adjacency. The skeleton uses connection
probability $0.45$, giving $13$ of the $30$ possible cross-client
couplings. The attention scoring and value scales are chosen so that the
generator's attention is strongly state-dependent and the cross-client
information gap is large (own-history oracle $1.6\times$ worse than the
full oracle). This is the \emph{realizable} benchmark and therefore the one dataset
where entrywise, time-resolved comparisons between the learned and true
operators are fully meaningful (Figure~\ref{fig:interpretability}); the
ground-truth Jacobian is computed analytically from the generator at every
test step.

\paragraph{Lorenz-96.}
The third system is the textbook Lorenz-96 chaotic dynamical system on the
standard ring of $N = 12$ sites,
\[
\frac{dx_i}{dt} = (x_{i+1} - x_{i-2})\,x_{i-1} - x_i + F,
\]
for $i = 1,\dots,12$, with forcing $F=8$ (the classic chaotic regime),
integrated with a fourth-order Runge-Kutta scheme at $dt = 0.05$ after a
burn-in period. The ring is vertically partitioned so that each client
owns two of the twelve sites, hence every advective interaction couples
two clients.
Unlike the Nonlinear SSM, this generator is \emph{not} a GAT; it is the
non-realizable physics benchmark, showing that the framework's structure
recovery is not an artifact of analyzing GAT-generated data with a GAT.

\subsection{Experimental Setup for the Synthetic Experiments}
\label{app:synth-setup}

\textbf{Proprietary models at the clients.} In the synthetic pipeline each client's
proprietary model is an extended Kalman filter whose assumed local dynamics
are deliberately coupling-blind: $f_m(h) = \tanh(\phi_m h)$ with client-specific gains,
paired with the true measurement function. The EKF supplies the local state
estimates $(\hat h_t^m)_c$ that enter federated training; the framework
treats it as a black box (cf.\ Appendix~\ref{app:algorithms}).

\textbf{Our model.} Each client's augmentation module $\Delta_m$ is an
LSTM over the window of its own raw observations, and the server is a
one-layer graph attention network masked by the given adjacency
(Appendix~\ref{app:algorithms}).

\textbf{Training.} The server and the client augmentation modules are
trained jointly with Adam (server learning rate $5\cdot10^{-4}$,
clients $10^{-3}$) for up to $300$ epochs with $1200$ subsampled
supervision time steps per epoch, with early stopping and per-client
freezing when a client's alignment loss plateaus. All results
are computed on the held-out test window and averaged over five training
seeds.

\textbf{Centralized oracle.} The Centralized Oracle is a
filter with access to \emph{all} clients' observations and the true
dynamics: the exact Kalman filter for the Linear SSM and the oracle EKF
for the two nonlinear systems. It is the best achievable reference, not a
trained competitor.

\section{Real-World Dataset Details}
\label{app:realworld}

All three real-world benchmarks are used in the same vertical-federated
construction: the sensor columns of one shared timeline are split by
physical subsystem, every client sees all timestamps but only its own
sensors, and no raw observations are ever exchanged. For each client $m$,
every column is z-scored by its \emph{nominal} (anomaly-free) mean and
standard deviation, and a rank-$p_m$ truncated SVD of the z-scored nominal
block defines a fixed linear decoder $C_{mm} \in \mathbb{R}^{d_m \times
p_m}$; latent states are $h_t^m = C_{mm}^\top y_t^m$ with $p_m = 3$ for
every client. This preprocessing is applied identically to our method and
all baselines; only the centralized reference instead applies a single
global SVD of equal total rank ($M \cdot p_m$) to the concatenated
observations of all clients.

\paragraph{\textbf{HAI}}
The HAI benchmark \cite{github_HAI} (HIL-based Augmented Industrial
control system security dataset, version 23.05) is collected from a
physical testbed combining a \textbf{boiler process (P1)}, a
\textbf{turbine process (P2)}, and a \textbf{water-treatment process
(P3)}, interconnected through a hardware-in-the-loop (HIL) simulator that
emulates thermal-power generation. The dataset provides $86$
measurement/actuation points sampled once per second. We treat the three
physical processes P1, P2, and P3 as clients (raw dimensions $37/24/7$). Three regimes of the official release are used in strictly disjoint roles:
the attack-free training stream is used for all training and calibration;
the first test stream supplies the attack window on which each client's
proprietary model is pre-trained on its own sensors alone, as an
independent unit (Appendix~\ref{app:add-real}); and the second test stream
($38$ labeled attacks) is the evaluation stream for
root-cause analysis. Because the proprietary pre-training window comes from
the first test stream and evaluation runs exclusively on the second, the
evaluation is leakage-free by construction and no events are excluded.
Ground truth per event (attack window and attacked-client set) is derived
from the official attack documentation.

\textit{\textbf{TEP}}
The Tennessee Eastman Process \cite{TEP} is the standard chemical-plant
benchmark; we use the Rieth et al.\ (2017) release (Harvard Dataverse), a
simulated plant with a reactor, condenser, separator, compressor, purge,
and stripper, providing $52$ variables ($41$ process measurements and $11$
manipulated variables) at 3-minute samples. The fault-free portion is used for
training and calibration. The faulty portion covers the disturbances with
a documented physical origin (IDV~1 to IDV~15), giving $300$ evaluation
events in total, each with a single root-cause label: the subsystem where
the fault physically originates. The $52$ variables are split into $M=5$ clients by physical unit: C1 Feeds,
C2 Reactor, C3 Separator+Condenser+Compressor, C4 Purge, and C5 Stripper.

\textit{\textbf{SMD.}}
The Server Machine Dataset \cite{SMD}, released with OmniAnomaly, contains
telemetry from $28$ server machines, each with the same $38$ anonymized
metrics (CPU, memory, disk, network, \dots) sampled once per minute, with
an anomaly-free training half and a test half with labeled anomalies. The $38$ metrics are split into $M=5$ clients of contiguous metrics.
Ground truth comprises $327$ labeled anomaly events with multi-label
attacked-client sets. The cross-client dependence in SMD is real but
symmetric (common-cause).
SMD therefore serves as the multi-label localization benchmark; strong
directional-attribution margins are not expected there, and that
expectation is part of the analysis in Appendix~\ref{app:add-real}.

\section{Additional Experiment Details}
\label{app:additional}

\subsection{Additional Results on Synthetic Experiments}
\label{app:add-synth}

\subsubsection{Toy example: the $M=3$ case}
\label{app:synth-m3}

\begin{figure}[htbp]
\centering
\includegraphics[width=\columnwidth]{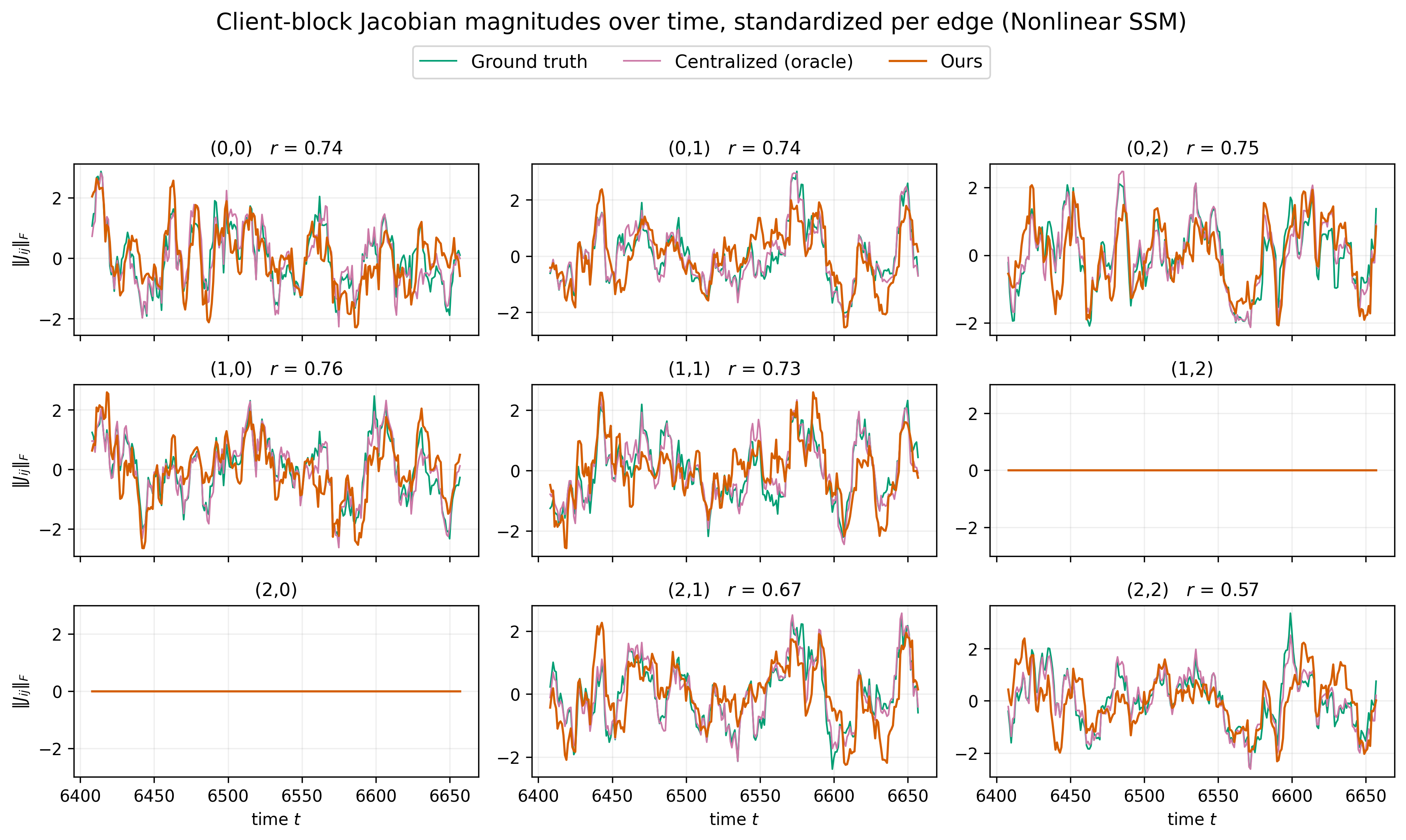}
\caption{Toy example ($M=3$ nonlinear SSM, same oracle and server GAT as Section~\ref{sec:Convergence}):
edge-by-edge temporal tracking of the Jacobian magnitudes $\|J_{mn}(t)\|$
(standardized), learned server vs.\ ground truth, over the $3\times3$ grid
of client pairs.}
\label{fig:app-m3-jac}
\end{figure}

\begin{figure}[htbp]
\centering
\includegraphics[width=\columnwidth]{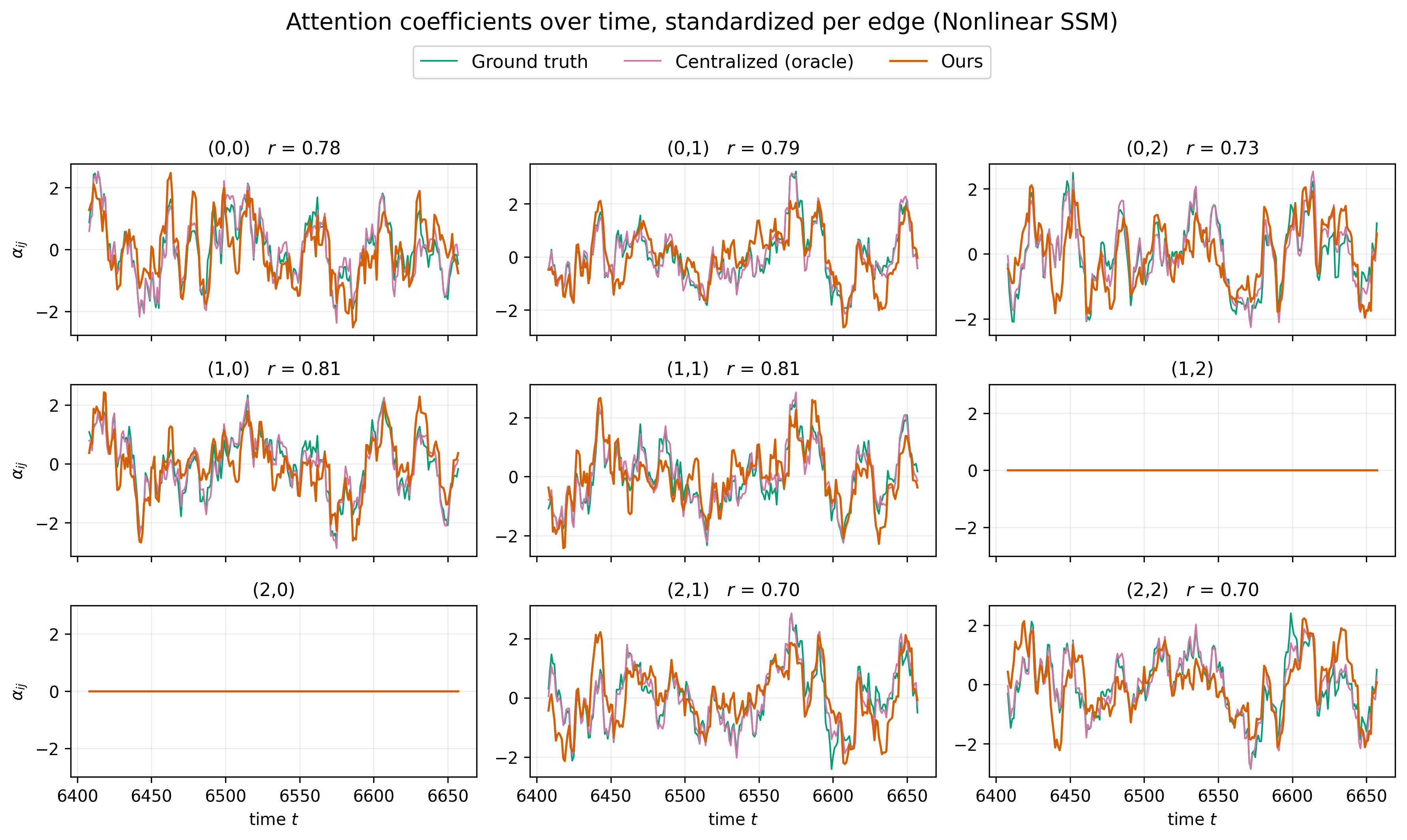}
\caption{Toy example ($M=3$): edge-by-edge temporal tracking of the
attention coefficients $\alpha_{mn}(t)$ (standardized) over the $3\times3$
grid of client pairs.}
\label{fig:app-m3-alpha}
\end{figure}

\begin{figure}[htbp]
\centering
\includegraphics[width=0.6\columnwidth]{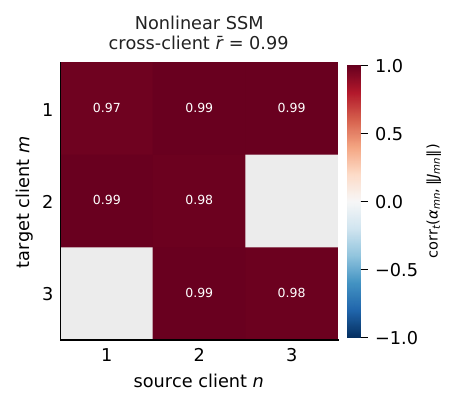}
\caption{Toy example ($M=3$): per-edge temporal correlation between
attention coefficients and Jacobian magnitudes.}
\label{fig:app-m3-f4}
\end{figure}

Alongside the $M=6$ benchmarks, we use a small $M=3$ toy configuration of
the Nonlinear SSM, for the
theoretical-validation figures of the main text
(Figures~\ref{fig:training}, \ref{fig:client_residuals}, and
\ref{fig:interpretability}(a) and (b)). In this toy example, the oracle and the server
employ the same GAT architecture and activation function. With three clients, the attention and Jacobian operators are $3\times3$
grids of client pairs, which can be inspected edge by edge.
Figures~\ref{fig:app-m3-jac} and \ref{fig:app-m3-alpha} give this
time-resolved view behind Figure~\ref{fig:interpretability}: for every
client pair $(m,n)$, the standardized traces of the learned server's
Jacobian magnitude $\|J_{mn}(t)\|$ and attention coefficient
$\alpha_{mn}(t)$ move together with their ground-truth counterparts over
the test window. Figure~\ref{fig:app-m3-f4} summarizes the same content as
the per-edge temporal correlation between attention and Jacobian
magnitude, the empirical counterpart of
Proposition~\ref{prop:jacobian_gat} in this realizable setting.

\subsubsection{State dependence: the operator moves with the dynamics}
\begin{figure}[htbp]
\centering
\includegraphics[width=\columnwidth]{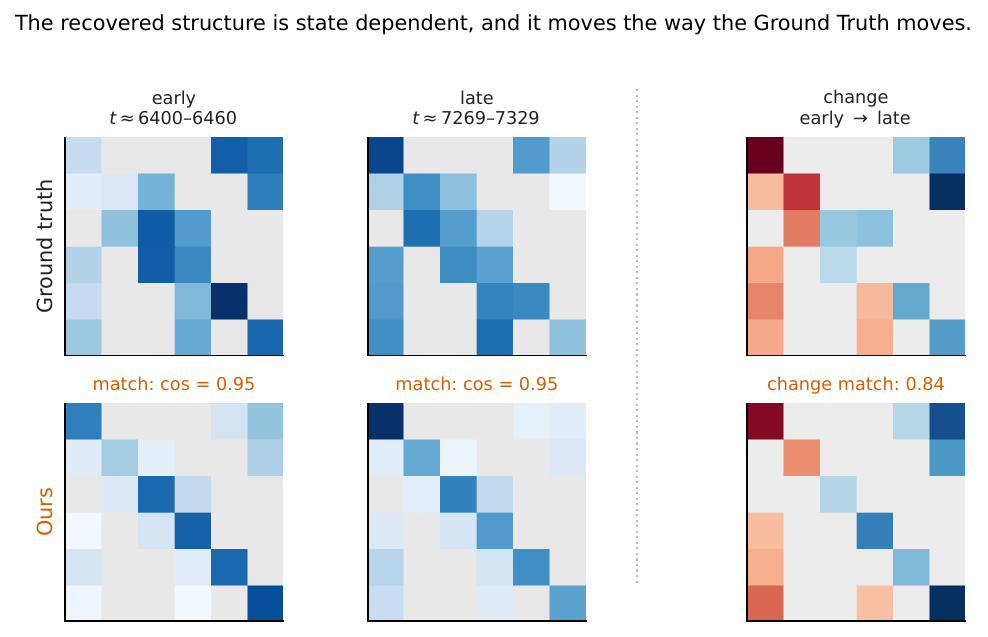}
\caption{State dependence of the learned operator (Nonlinear SSM). Left of
the dotted line: window-averaged $\|J_{mn}\|$ in an early and a late window,
ground truth (top) vs.\ ours (bottom); each map matches the truth
(match cosine annotated). Right: the between-window \emph{change} map: the
learned operator changes in the same places and directions as the truth
(change-match annotated). Gray = no edge; color scales are per row.}
\label{fig:app-statedep}
\end{figure}

Every baseline learns one \emph{fixed} influence map: a single matrix
meant to describe the system at all times. But who influences whom changes
with the state, and our server's operator is a function of the state. To
show it changes \emph{correctly}, we take two windows of the test data of
the Nonlinear SSM, one early and one late, and average the ground-truth and
estimated Jacobians in each. Two checks follow
(Figure~\ref{fig:app-statedep}): (i) each window-averaged map matches the
truth, and (ii), the stronger claim, the \emph{change} between the
windows matches in location and direction (cosine values annotated in the
figure; stable across seeds). A fixed-matrix method cannot produce this figure: its two maps are
identical by definition and its change map is exactly zero.

\subsection{Additional Results on Real-World Experiments}
\label{app:add-real}

\subsubsection{Proprietary client models}
To keep the setting realistic and the mechanism identifiable, each client's
proprietary model $f_m$ is a small recurrent predictor (single-layer LSTM,
hidden size $32$) over its own latent history (window $S=8$), pre-trained
on the client's own sensor data alone, with no access to cross-client
information, and then frozen during federated training. 

Each $f_m$ is trained with Adam (learning rate $10^{-3}$, batch size
$512$) and subsequently frozen. This design explains the spread of the
loss reductions in Table~\ref{tab:mse}: on HAI the proprietary models are
missing exactly the nominal cross-client structure, so the federation
recovers a large fraction; on TEP the local nominal dynamics are largely
learnable from each unit's own sensors, leaving little loss headroom, so
TEP's evidence is attribution rather than recovery; SMD is intermediate.

\subsubsection{Federated training and centralized baseline}
During federated training each client learns an augmentation module
$\Delta_m$: a single-layer LSTM (hidden size $64$) over the raw sensor
window producing a latent
correction that feeds the frozen proprietary predictor. The server
consumes the clients' latent histories through per-client LSTM encoders
(hidden size $64$), since real systems exhibit multi-step temporal
dependencies, followed by a single-layer graph attention network over a
fully connected candidate graph (no prior graph knowledge is assumed on
real data). Only the server parameters and the $\Delta_m$ are optimized
(Adam, learning rate $10^{-3}$, batch size $512$, up to $200$ epochs) on
the nominal stream with an 80/20 chronological split, early stopping on
the validation alignment loss. The centralized reference
uses the same LSTM+GAT architecture on the global SVD latent space of
equal total rank and is trained with the same optimizer, split, and
stopping rule; it models global interdependencies under full data access
and serves as the upper-bound reference.

\subsubsection{Root-cause evaluation protocol (AC@$k$)}
\label{app:ack-protocol}
Every method is scored with identical rules; the \emph{only} per-method
ingredient is its interdependency operator.

\textit{\textbf{Problem setup.}} Let there be $M$ clients. For each client $j$, let
$z_j(t) \in \{0,1\}$ denote whether client $j$ is anomalous at time $t$.
In practice $z_j(t)$ is produced by a shared, model-free detector: the
deviation signal $\mathrm{dev}_j(t) = \|(x_t^j - \mu_j)/\sigma_j\|_2$
(nominal z-scores of the client's raw sensors) is compared against a
threshold calibrated on nominal data, identically for every method. We
define the first anomaly arrival time for client $j$ as
\[
t_j = \min\{t : z_j(t) = 1\}.
\]
If client $j$ is never anomalous during the incident window we set
$z_j = 0$; otherwise $z_j = 1$, and $t_j$ is the first time at which the
anomaly appears at client $j$. We assume that root-cause arrivals follow
a Poisson process, so simultaneous root-cause arrivals at multiple
clients occur with probability zero and each incident has a unique
earliest arrival.

\textit{\textbf{Temporal propagation score.}} Let $J(t)$ denote the learned
time-varying interdependency operator; the block $J_{ij}(t)$ represents
the influence from source client $j$ to target client $i$ at time $t$.
For the nonlinear transition $h_t = f(h_{t-1})$, the Jacobian
$J(t) = \partial f(h_{t-1})/\partial h_{t-1}$ characterizes the local
propagation of perturbations,
\[
\delta h_t \approx J(t)\, \delta h_{t-1},
\]
so an $\ell$-step perturbation propagates locally as
\[
\delta h_{t+\ell} \approx J(t+\ell)\, J(t+\ell-1) \cdots J(t+1)\, \delta h_t .
\]
Therefore, the block $\big(J(t_i) \cdots J(t_j+1)\big)_{ij}$ represents
the influence of a perturbation at client $j$ on client $i$ across the
elapsed lag $t_i - t_j$; for a static operator $A$ this product reduces to
the matrix power $A^{t_i - t_j}$. We define the root-cause score for
candidate client $j$ as

\begin{multline*}
s_j = \sum_{i=1}^{M} z_j z_i\, \mathbf{1}(t_j \le t_i)\,
\beta^{\,t_i - t_j} \\
\times \big\| \big( J(t_i)\, J(t_i-1) \cdots J(t_j+1) \big)_{ij}
\big\|_F ,
\end{multline*}
where $0 < \beta \le 1$ is a temporal discount factor ($\beta = 1$ in the
main tables). Clients are ranked in descending order of
$s_j$; the top-ranked clients are considered the most likely root causes.
For our method, $J(t)$ is the server's Jacobian, whose
entry $J_{ij}(t)$ is the norm of the derivative of client $i$'s decoded
prediction with respect to client $j$'s input window.

\textit{\textbf{Explanation of the score components.}} The term $z_j z_i$ ensures
that only anomalous clients contribute to the score: a candidate root
cause must itself be anomalous, and it receives evidence only from other
anomalous clients. The indicator $\mathbf{1}(t_j \le t_i)$ enforces
temporal ordering: client $j$ can only receive credit for explaining
client $i$ if the anomaly at $j$ appears no later than the anomaly at
$i$. The factor $\beta^{\,t_i - t_j}$ discounts delayed propagation. The
term $\big(J(t_i) \cdots J(t_j+1)\big)_{ij}$ captures multi-step
propagation through the learned dynamics along the actual trajectory,
which allows the score to account for indirect paths such as
$1 \to 2 \to 3$. The Frobenius norm measures the strength of the
propagated influence from client $j$ to client $i$. Overall, $s_j$ is
large when client $j$ becomes anomalous early and strongly explains
later anomalies through the learned dynamics.

For each incident $q$, let $R^\star_q$
denote the set of ground-truth root-cause clients  and let $\mathrm{Top}\text{-}k(q)$
denote the set of top-$k$ clients ranked by the temporal score $s_j$.
Then
\[
\mathrm{AC@}k = \frac{1}{Q} \sum_{q=1}^{Q}
\mathbf{1}\big( R^\star_q \cap \mathrm{Top}\text{-}k(q) \neq \emptyset \big),
\]
where $Q$ is the total number of incidents.

\subsubsection{Fixed-operator controls for the RCA score}
\label{app:rca-controls}
Because the propagation score of Appendix~\ref{app:ack-protocol}
propagates the anomaly indicators through the learned operator $J(t)$
itself, we check how much of the RCA performance is due to anomaly timing
alone, keeping the detector, the arrival times, and the ranking rule
unchanged. Table~\ref{tab:rca-controls} replaces $J(t)$ by two fixed
reference operators that do not depend on any learned model: a fixed
identity matrix, which keeps only the intra-client terms and therefore
scores on timing alone, and a fixed all-ones matrix, which treats every
cross-client interaction as equally strong and is the best-case fixed
operator on a fully connected candidate graph.

\begin{table}[htbp]
\centering
\caption{RCA with fixed reference operators in place of the learned
$J(t)$, under the same detector, timing, and ranking as Table~\ref{tab:rca}.}
\label{tab:rca-controls}
\setlength{\tabcolsep}{2pt}
\renewcommand{\arraystretch}{1}
{\small
\begin{tabular}{lcc|cc|cc}
\hline
& \multicolumn{2}{c|}{\textbf{HAI}}
& \multicolumn{2}{c|}{\textbf{TEP}}
& \multicolumn{2}{c}{\textbf{SMD}} \\
\textbf{Operator}
& \textbf{AC@1} & \textbf{AC@2}
& \textbf{AC@1} & \textbf{AC@2}
& \textbf{AC@1} & \textbf{AC@2} \\
\hline
Fixed identity $J$ & 0.43 & 0.77 & 0.20 & 0.40 & 0.44 & 0.64 \\
Fixed all-ones $J$ & 0.56 & 0.85 & 0.40 & 0.63 & 0.52 & 0.70 \\
\textbf{Ours, learned $J(t)$}
& \textbf{0.76} & \textbf{0.94}
& \textbf{0.45} & \textbf{0.65}
& \textbf{0.53} & \textbf{0.70} \\
\hline
\end{tabular}}
\end{table}

With the identity operator, performance drops substantially on HAI and
TEP, so the temporal terms alone do not explain the RCA performance
without knowledge of the cross-client interdependencies. The all-ones
operator is the stronger fixed baseline, and even this baseline falls
short of the learned $J(t)$ on HAI and TEP. The remaining differences
follow the dataset mechanisms discussed next: on SMD, where clients
become abnormal together with no clear ordering, the learned and the
all-ones operators are nearly identical.

\subsubsection{Why the three datasets behave differently}
Our method outperforms every baseline on all three datasets
(Table~\ref{tab:rca}), but the margins differ, and the differences are
mechanistically informative. The margin is largest on \textbf{HAI}: the
same trained model also delivers the largest loss reduction in
Table~\ref{tab:mse}, so the server is genuinely using cross-client
information rather than each client's own signal; once that
coupling is captured, ranking the true root first follows directly. On
\textbf{TEP}, the units are physically connected, so a fault originates in
one unit and spreads downstream; once it has propagated, ``who deviated
most'' no longer reveals where it began. Our method roughly doubles the
baselines because it uses the \emph{timing}: the time-varying operator
links the origin's early deviation to the later deviations downstream. The
margin is smallest on \textbf{SMD}, whose clients are coupled by a common
cause: they become abnormal together with no clear ordering of who reacted
first. Since our advantage comes from reading that ordering, there is
little signal to draw on, and the margin stays moderate, consistent
with SMD's role as the multi-label localization benchmark
(Appendix~\ref{app:realworld}).

\subsection{Additional Results on the Ablation Studies}
\label{app:add-ablation}

\subsubsection{Robustness: construction and loss-side results}
The random proprietary models of Table~\ref{tab:robustness} are frozen
randomly initialized networks rather than informative ones. In the subset sweep, clients are corrupted cumulatively with the same
mechanism while the rest stay healthy.

On the AC@k side, the sweep separates two regimes by whether the
corrupted clients contain the root causes. On TEP and SMD, attribution is
robust: SMD stays within $0.03$ of its uncorrupted value across the whole
sweep and TEP declines only mildly. HAI probes the hardest case: the
ground-truth roots concentrate on a single client, and the sweep corrupts
exactly that client first. The cost appears precisely where expected, at
AC@1: with the root client's own signal replaced by a meaningless one,
the server ranks the true root first less often. AC@2, however, holds
throughout the subset sweep: even when the root
client itself is corrupted, the true root remains within the top two, a
one-position degradation rather than a collapse. In all cases AC@$k$ stays
above every baseline of Table~\ref{tab:rca} at every corruption level.

\subsubsection{Graph misspecification: mechanism}
The perturbation of Table~\ref{tab:adjacency} is a density-matched
pair-swap: a fraction $\rho$ of the true edges is removed and, for each, a
spurious edge is inserted on a true structural zero, so the edge density
always matches the true graph and the corruption level is the only moving
variable; each level is averaged over 3 random draws of the corrupted edge
set, and the sweep is anchored at $\rho = 0$, which reproduces the
main-table model exactly. The two deterministic extremes separate the two
corruption mechanisms. \emph{Insertions}: the fully connected
graph (insertions only, which is also the topology the real-world experiments
run on) stays at or above the $\rho = 30\%$ level despite carrying more
wrong edges, because the GAT assigns inserted spurious edges
systematically less attention mass than true edges: it identifies and
down-weights the wrong edges it was given, ranking them below every true
edge. \emph{Deletions}: a masked edge carries exactly zero attention and zero Jacobian by construction, so a deleted true edge can never be recovered; restoring it would amount to graph discovery, which the framework deliberately does not perform. Practically, this asymmetry means a fully connected
graph is the safe default when no prior graph knowledge is available,
which is the choice made in all real-world experiments.

\subsubsection{Server architecture}
An informative interdependency operator must place influence on real
couplings only; Table~\ref{tab:server} therefore reports, alongside the
prediction loss, the fraction of off-diagonal influence mass assigned to
absent edges. The graph-blind alternatives are a one-layer Transformer
encoder over the $M$ client tokens (unmasked attention), a bidirectional
LSTM over the client axis, and an MLP on the flattened joint state; all
are given $4\times$ to $10\times$ the GAT's parameter count, which is
generous to the alternatives.

The alternatives match the GAT's prediction loss only on the Linear SSM
and fall behind on the two nonlinear systems, and every one of them
places a substantial share of its inter-client influence on couplings
that do not exist; in a root-cause reading, these are false influence
paths, and an operator with a large fraction of its mass on false edges
is uninformative for attribution regardless of its loss. The GAT assigns
absent edges exactly zero influence by construction, so no post-hoc
thresholding of a contaminated matrix is ever needed, and it achieves
the best prediction loss at the lowest parameter count.

\subsection{Scalability}
\label{appendix:scalability}

We measure how the framework scales along its three size parameters, namely
the observation dimension $d_m$, the latent state dimension $p_m$, and the
number of clients $M$, starting from the $M{=}3$, $p_m{=}2$, $d_m{=}16$
base system and changing one parameter
at a time. Per round, each client communicates only its $p_m$-dimensional
state vector, never its $d_m$-dimensional observations, so the
communication cost grows linearly in $M$ and $p_m$ and does not depend on
$d_m$ at all, while the number of server parameters is independent of the
data size. Table~\ref{tab:scalability} reports the server loss $L_s$ on
the test window (mean $\pm$ s.d.\ over 3 seeds).

\begin{table}[htbp]
\centering
\caption{Scalability: server loss on the test window (mean $\pm$ s.d.\
over 3 seeds; base $M{=}3$, $p_m{=}2$, $d_m{=}16$) and wall-clock time of
the Jacobian read-out per time step. The $M$ rows keep the cross-client graph
density fixed (Table~\ref{tab:scal-graphs}).}
\label{tab:scalability}
\setlength{\tabcolsep}{4pt}
\renewcommand{\arraystretch}{1.1}
{\small
\begin{tabular}{c c c c}
\toprule
\textbf{Scale w.r.t.} & \textbf{Value} & \textbf{Server loss $L_s$} & \textbf{Read-out (s/step)} \\
\midrule
{} & 8   & $0.0022 \pm 0.0001$ & $0.0003$ \\
{} & 16  & $0.0023 \pm 0.0001$ & $0.0003$ \\
Observation dim.\ $d_m$ & 32  & $0.0025 \pm 0.0002$ & $0.0003$ \\
{} & 64  & $0.0028 \pm 0.0003$ & $0.0003$ \\
{} & 128 & $0.0031 \pm 0.0004$ & $0.0003$ \\
\midrule
{} & 2 & $0.0023 \pm 0.0001$ & $0.0003$ \\
State dim.\ $p_m$ & 4 & $0.0015 \pm 0.0001$ & $0.0005$ \\
{} & 8 & $0.0017 \pm 0.0000$ & $0.0010$ \\
\midrule
{} & 3   & $0.0023 \pm 0.0001$ & $0.0003$ \\
{} & 4   & $0.0023 \pm 0.0001$ & $0.0004$ \\
{} & 8   & $0.0021 \pm 0.0004$ & $0.0007$ \\
Number of clients $M$ & 16 & $0.0019 \pm 0.0000$ & $0.0018$ \\
{} & 32  & $0.0016 \pm 0.0001$ & $0.0036$ \\
{} & 64  & $0.0015 \pm 0.0001$ & $0.0086$ \\
{} & 128 & $0.0013 \pm 0.0001$ & $0.0267$ \\
{} & 256 & $0.0014 \pm 0.0001$ & $0.1808$ \\
\bottomrule
\end{tabular}}
\end{table}

\textit{\textbf{Performance is independent of the observation dimension.}}
Server loss is essentially flat from $d_m = 8$ to $128$.
The reason is structural:
the server never sees the observations (clients send only their
$p_m$-dimensional states), so $d_m$ affects nothing beyond how well each
client estimates its own state from its own measurements. Once that local
estimate is good, giving a client more sensors changes nothing about the
server's task.

\textit{\textbf{No systematic cost with the state dimension.}} The loss stays
within one narrow band as the state dimension increases. This is the expected behaviour of
the architecture: the server's attention weights are $p_m \times p_m$
matrices, so its capacity grows together with the state it must track,
while the message each client sends grows only linearly in $p_m$. Nothing
in the table suggests that larger client states are systematically harder
for the federation.

\textit{\textbf{Client count: stable loss, linear communication, constant
server size.}} From $M{=}3$ to $256$ the server loss does not degrade
($0.0023 \to 0.0014$): a single shared attention model serves the
clients, the amount of communication grows only in proportion to $M$,
and the number of server parameters does not grow at all. The numbers
even drift slightly \emph{downward} as $M$ grows, and it is tempting to
read that as the federation getting \emph{better} with more clients. It
is not: because the graph density is held fixed (the same fraction of
client pairs is connected at every $M$), each client has more neighbours
in a larger federation. The attention mechanism forms each prediction by
averaging over a client's neighbours, and an average over many sources
is steadier than any single source, so as $M$ grows the signals the
server must predict become smoother and easier to predict. The loss
shrinks because the task became gentler, not because the model became
stronger. The honest claim is therefore stability: the server loss does
not degrade out to $M{=}256$, at linear communication cost and constant
model size.

\textit{\textbf{Jacobian read-out.}} The interdependency operator is read
out once per time step after training. Its cost (Table~\ref{tab:scalability})
is independent of $d_m$, grows mildly with $p_m$, and grows faster than
linearly only with $M$: below $0.01$\,s per step up to $M{=}64$,
$0.027$\,s at $M{=}128$, and $0.18$\,s at $M{=}256$, i.e., about $3$
minutes for a $1000$-step window. Being a post-training operation, this
cost is paid once per analysis window and does not enter the federated
training or the communication rounds.

\textit{\textbf{Realized graphs of the $M$ sweep.}}
Table~\ref{tab:scal-graphs} lists the realized cross-edge statistics: the
density is preserved at every $M$, so the per-client neighbourhood grows
in proportion to $M$, the regime of a realistic decentralized network
in which a client's number of couplings grows with the network.

\begin{table}[htbp]
\centering
\caption{Realized adjacency statistics of the $M$ sweep: the cross-edge
density is $2/3$ at every $M$ (up to integer rounding of the edge count).}
\label{tab:scal-graphs}
\setlength{\tabcolsep}{6pt}
\renewcommand{\arraystretch}{1.1}
{\small
\begin{tabular}{c c c c}
\toprule
$M$ & \textbf{Cross edges} & \textbf{Possible} & \textbf{Density} \\
\midrule
3  & 4   & 6   & $0.667$ \\
4  & 8   & 12  & $0.667$ \\
8  & 37  & 56  & $0.661$ \\
16 & 160 & 240 & $0.667$ \\
32 & 661 & 992 & $0.666$ \\
64 & 2688 & 4032 & $0.667$ \\
128 & 10837 & 16256 & $0.667$ \\
256 & 43520 & 65280 & $0.667$ \\

\bottomrule
\end{tabular}}
\end{table}


\subsection{Discussion on Privacy Analysis}
\label{appendix:privacy}

We analyze the privacy of both communication directions (server-to-client and client-to-server)  in the proposed
framework. First, clients may perturb their estimated latent states before
transmitting them to the server. This provides privacy against an
honest-but-curious coordinating server. Second, the server may clip and
perturb the partial gradients returned to the clients. This protects the
contribution of one client from being inferred by another client through
the server-to-client communication channel.

These mechanisms correspond to different threat models and may be used
independently. Client-to-server perturbation is appropriate when the server
is not trusted with the clients' latent representations. Server-to-client
perturbation is appropriate when the server is trusted, but individual
clients may attempt to infer information about other clients from the
gradients they receive. The two mechanisms may also be combined when
privacy is required in both communication directions.

We begin by introducing the definitions required for the analysis. Readers
are referred to \cite{DP_book,RDP,DPSGD} for detailed discussions of
differential privacy, R\'enyi differential privacy, and privacy-preserving
gradient perturbation.

\subsubsection{Preliminaries}

\begin{definition}[\textbf{Differential Privacy}]
A randomized mechanism $\mathcal{M}$ satisfies
$(\varepsilon,\delta)$-differential privacy if, for every measurable
subset $\mathcal{S}$ of the output space and every pair of neighboring
datasets $D$ and $D'$,
\begin{equation}
\Pr\left[\mathcal{M}(D)\in\mathcal{S}\right]
\leq
e^{\varepsilon}
\Pr\left[\mathcal{M}(D')\in\mathcal{S}\right]
+
\delta.
\end{equation}
\end{definition}

\begin{definition}[\textbf{Neighboring Datasets}]
\label{definition:privacy_adjacency}
A protected data unit may represent one observation, one temporal window,
or one complete trajectory.

For client-to-server communication, two local datasets (raw observations) $D_m$ and $D_m'$
are neighboring, denoted by $D_m\simeq D_m'$, if they differ in at most
one protected data unit belonging to client $m$.

For server-to-client communication, let
$D=(D_1,\ldots,D_M)$ denote the collection of client datasets (state estimates) available
to the server. Two global datasets $D$ and $D'$ are neighboring with
respect to client $n$, denoted by $D\simeq_n D'$, if they differ in at
most one protected data unit belonging to client $n$ and are identical
for all other clients.
\end{definition}

The resulting guarantee is \textit{window-level differential privacy} when one
temporal window is treated as the protected unit, and \textit{trajectory-level
differential privacy} when one complete trajectory is treated as the
protected unit.

\begin{definition}[\textbf{$\ell_2$-Sensitivity}]
The $\ell_2$-sensitivity of a function
$f:\mathcal{D}\rightarrow\mathbb{R}^d$ is
\begin{equation}
\Delta_2(f)
=
\sup_{D\simeq D'}
\left\|
f(D)-f(D')
\right\|_2.
\end{equation}
For matrix-valued functions, the corresponding sensitivity is measured
using the Frobenius norm, which is equivalent to the vector
$\ell_2$-norm after vectorization.
\end{definition}

\begin{definition}[\textbf{Gaussian Mechanism}]
Let $f:\mathcal{D}\rightarrow\mathbb{R}^d$ have $\ell_2$-sensitivity
$\Delta_2(f)$. The Gaussian mechanism releases
\begin{equation}
\mathcal{M}(D)
=
f(D)+Z,
\qquad
Z\sim\mathcal{N}(0,\sigma^2 I_d).
\end{equation}
For $\varepsilon\in(0,1)$, the classical Gaussian mechanism satisfies
$(\varepsilon,\delta)$-differential privacy if
\begin{equation}
\sigma
\geq
\frac{
\Delta_2(f)
\sqrt{2\log(1.25/\delta)}
}{
\varepsilon
}.
\end{equation}
\end{definition}

\begin{definition}[\textbf{Sequential Composition}]
\label{definition:sequential_composition}
Suppose that mechanisms $\mathcal{M}_1,\ldots,\mathcal{M}_R$ satisfy
$(\varepsilon_r,\delta_r)$-differential privacy for
$r=1,\ldots,R$. Their adaptive sequential composition satisfies
\begin{equation}
\left(
\sum_{r=1}^{R}\varepsilon_r,
\sum_{r=1}^{R}\delta_r
\right)
\text{-differential privacy}.
\end{equation}
\end{definition}

\begin{definition}[\textbf{Clipping Function}]
\label{definition:clipping_function}
Given a matrix $G$ and a clipping threshold $C_g>0$, define
\begin{equation}
\operatorname{Clip}(G,C_g)
=
G
\min\left\{
1,
\frac{C_g}{\|G\|_F}
\right\}.
\end{equation}
The clipped matrix satisfies
$\|\operatorname{Clip}(G,C_g)\|_F\leq C_g$.
\end{definition}

\begin{definition}[\textbf{R\'enyi Differential Privacy}]
A randomized mechanism $\mathcal{M}$ satisfies
$(\alpha,\varepsilon_{\alpha})$-R\'enyi differential privacy for
$\alpha>1$ if, for every pair of neighboring datasets $D$ and $D'$,
\begin{equation}
D_{\alpha}
\left(
\mathcal{M}(D)
\|
\mathcal{M}(D')
\right)
\leq
\varepsilon_{\alpha},
\end{equation}
where $D_{\alpha}(\cdot\|\cdot)$ denotes the R\'enyi divergence of order
$\alpha$.
\end{definition}

If a Gaussian mechanism has sensitivity $\Delta_2$ and noise standard
deviation $\sigma$, then it satisfies
\begin{equation}
\left(
\alpha,
\frac{\alpha\Delta_2^2}{2\sigma^2}
\right)
\text{-R\'enyi differential privacy}.
\end{equation}

R\'enyi privacy parameters compose additively under adaptive sequential
composition. Therefore, if mechanism $\mathcal{M}_r$ satisfies
$(\alpha,\varepsilon_{\alpha}^{(r)})$-R\'enyi differential privacy, then
the complete transcript satisfies
\begin{equation}
\left(
\alpha,
\sum_{r=1}^{R}
\varepsilon_{\alpha}^{(r)}
\right)
\text{-R\'enyi differential privacy}.
\end{equation}

An $(\alpha,\varepsilon_{\alpha})$-R\'enyi private mechanism also
satisfies $(\varepsilon,\delta)$-differential privacy for any
$\delta\in(0,1)$, where
\begin{equation}
\varepsilon
=
\varepsilon_{\alpha}
+
\frac{\log(1/\delta)}{\alpha-1}.
\end{equation}

\subsubsection{Client-to-Server Privacy}

We first consider the setting in which we have a \textbf{\textit{honest-but-curious server}}. Each client perturbs the latent states transmitted to
the server. Unlike gradient clipping in the server-to-client mechanism,
we do not clip the latent states because directly rescaling them may
distort the temporal dynamics learned by the server. Instead, we bound
their sensitivity using assumptions on the local observations and the
client state-estimation map.

Let $\mathbf{Y}_m^{(r)}$ denote the observation block used by client $m$
during communication round $r$. This block may contain one temporal
window or one complete trajectory. Let
$\Phi_m^{(r)}$ denote the client state-estimation map, such that
\begin{equation}
\widehat{\mathbf{H}}_m^{(r)}
=
\Phi_m^{(r)}
\left(
\mathbf{Y}_m^{(r)}
\right).
\end{equation}

The map $\Phi_m^{(r)}$ includes the complete local computation used to
obtain the latent-state block, including any recurrent state updates.

\begin{assumption}[\textbf{Bounded Observation Blocks}]
\label{assumption:bounded_observation}
For every client $m$, communication round $r$, and admissible observation
block,
\begin{equation}
\left\|
\mathbf{Y}_m^{(r)}
\right\|_F
\leq
B_{y,m}.
\end{equation}
\end{assumption}

\begin{assumption}[\textbf{Lipschitz State-Estimation Map}]
\label{assumption:lipschitz_state_map}
For every client $m$ and communication round $r$, the state-estimation
map $\Phi_m^{(r)}$ is $L_{h,m}$-Lipschitz:
\begin{equation}
\left\|
\Phi_m^{(r)}(\mathbf{Y})
-
\Phi_m^{(r)}(\mathbf{Y}')
\right\|_F
\leq
L_{h,m}
\left\|
\mathbf{Y}-\mathbf{Y}'
\right\|_F.
\end{equation}
The bound is assumed to hold conditionally on any transcript released
before round $r$.
\end{assumption}

Assumption~\ref{assumption:bounded_observation} may be enforced through
known physical measurement ranges or data normalization.
Assumption~\ref{assumption:lipschitz_state_map} requires a uniform bound
on the sensitivity of the complete local state-estimation model. For a
neural state estimator, such a bound may be imposed through constraints
on the operator norms of its layers. Because the latent states are not
clipped, a finite global sensitivity guarantee requires assumptions of
this form.

\begin{lemma}[\textbf{Sensitivity of the Latent-State Block}]
\label{lemma:latent_sensitivity}
Under Assumptions~\ref{assumption:bounded_observation} and
\ref{assumption:lipschitz_state_map}, the latent-state query
$\Phi_m^{(r)}$ has Frobenius sensitivity
\begin{equation}
\Delta_{h,m}
\leq
2L_{h,m}B_{y,m}
\end{equation}
under replacement adjacency.
\end{lemma}

\begin{proof}
For neighboring observation blocks $\mathbf{Y}_m^{(r)}$ and
$\mathbf{Y}_m^{\prime(r)}$, Lipschitz continuity gives
\begin{equation}
\begin{aligned}
&
\left\|
\Phi_m^{(r)}
\left(
\mathbf{Y}_m^{(r)}
\right)
-
\Phi_m^{(r)}
\left(
\mathbf{Y}_m^{\prime(r)}
\right)
\right\|_F
\\
&\leq
L_{h,m}
\left\|
\mathbf{Y}_m^{(r)}
-
\mathbf{Y}_m^{\prime(r)}
\right\|_F
\\
&\leq
L_{h,m}
\left(
\left\|
\mathbf{Y}_m^{(r)}
\right\|_F
+
\left\|
\mathbf{Y}_m^{\prime(r)}
\right\|_F
\right)
\\
&\leq
2L_{h,m}B_{y,m}.
\end{aligned}
\end{equation}
\end{proof}

\paragraph{Private Latent-State Mechanism.}
At communication round $r$, client $m$ releases
\begin{equation}
\widetilde{\mathbf{H}}_m^{(r)}
=
\widehat{\mathbf{H}}_m^{(r)}
+
Z_{h,m}^{(r)},
\end{equation}
where
\begin{equation}
\operatorname{vec}
\left(
Z_{h,m}^{(r)}
\right)
\sim
\mathcal{N}
\left(
0,
\sigma_{h,m}^2 I_{p_h}
\right).
\end{equation}
Here, $p_h$ is the number of scalar elements in the released latent-state
block and $\sigma_{h,m}$ is the actual standard deviation of the noise
added to each element.

\begin{theorem}[\textbf{Client-to-Server Communication}]
\label{theorem:privacy_c2s_single}
At any communication round $r$, the mechanism by which client $m$
releases $\widetilde{\mathbf{H}}_m^{(r)}$ satisfies
$(\varepsilon_h,\delta_h)$-differential privacy with respect to its
protected observation block, provided that
\begin{equation}
\sigma_{h,m}
\geq
\frac{
2L_{h,m}B_{y,m}
\sqrt{2\log(1.25/\delta_h)}
}{
\varepsilon_h
},
\qquad
\varepsilon_h\in(0,1).
\end{equation}
\end{theorem}

\begin{proof}
By Lemma~\ref{lemma:latent_sensitivity}, the sensitivity of the released
latent-state block is at most $2L_{h,m}B_{y,m}$. The result follows by
applying the Gaussian mechanism.
\end{proof}

The preceding theorem provides a guarantee for one release. The complete
privacy guarantee must account for all latent-state blocks communicated
during training.

\begin{theorem}[\textbf{Privacy of the Client-to-Server Transcript}]
\label{theorem:privacy_c2s_transcript}
Suppose client $m$ releases $R_h$ independently perturbed latent-state
blocks using sensitivity bound $\Delta_{h,m}$ and noise standard
deviation $\sigma_{h,m}$. For every $\alpha>1$, the complete transcript
\begin{equation}
\mathcal{T}_{h,m}
=
\left(
\widetilde{\mathbf{H}}_m^{(1)},
\ldots,
\widetilde{\mathbf{H}}_m^{(R_h)}
\right)
\end{equation}
satisfies
\begin{equation}
\left(
\alpha,
\frac{
R_h\alpha\Delta_{h,m}^2
}{
2\sigma_{h,m}^2
}
\right)
\text{-R\'enyi differential privacy}.
\end{equation}
Consequently, for every $\delta_h\in(0,1)$, the transcript satisfies
$(\varepsilon_h,\delta_h)$-differential privacy, where
\begin{equation}
\varepsilon_h
=
\min_{\alpha>1}
\left\{
\frac{
R_h\alpha\Delta_{h,m}^2
}{
2\sigma_{h,m}^2
}
+
\frac{\log(1/\delta_h)}{\alpha-1}
\right\}.
\end{equation}
\end{theorem}

\begin{proof}
Each Gaussian release satisfies
\begin{equation}
\left(
\alpha,
\frac{
\alpha\Delta_{h,m}^2
}{
2\sigma_{h,m}^2
}
\right)
\text{-R\'enyi differential privacy}.
\end{equation}
R\'enyi privacy parameters compose additively under adaptive sequential
composition. Summing over the $R_h$ releases and converting the resulting
R\'enyi guarantee to $(\varepsilon_h,\delta_h)$-differential privacy
proves the result.
\end{proof}

\begin{corollary}[\textbf{Client-to-Server Noise Calibration}]
\label{corollary:c2s_noise_calibration}
For a target privacy budget $(\varepsilon_h,\delta_h)$ over $R_h$
latent-state releases, it is sufficient to choose
\begin{equation}
\sigma_{h,m}
\geq
\frac{
\Delta_{h,m}\sqrt{R_h}
}{
\sqrt{2}
\left(
\sqrt{\log(1/\delta_h)+\varepsilon_h}
-
\sqrt{\log(1/\delta_h)}
\right)
}.
\end{equation}
Using $\Delta_{h,m}\leq2L_{h,m}B_{y,m}$ gives
\begin{equation}
\sigma_{h,m}
\geq
\frac{
L_{h,m}B_{y,m}\sqrt{2R_h}
}{
\sqrt{\log(1/\delta_h)+\varepsilon_h}
-
\sqrt{\log(1/\delta_h)}
}.
\end{equation}
\end{corollary}

\begin{proof}
Let
\begin{equation}
c_h
=
\frac{
R_h\Delta_{h,m}^2
}{
2\sigma_{h,m}^2
}
\end{equation}
and let $L_h=\log(1/\delta_h)$. Minimizing the expression in
Theorem~\ref{theorem:privacy_c2s_transcript} over $\alpha>1$ gives
\begin{equation}
\varepsilon_h
=
c_h+2\sqrt{c_hL_h}.
\end{equation}
Therefore,
\begin{equation}
\sqrt{c_h}
=
\sqrt{L_h+\varepsilon_h}
-
\sqrt{L_h}.
\end{equation}
Substituting the definition of $c_h$ and rearranging gives the stated
noise calibration.
\end{proof}

\begin{corollary}[\textbf{Privacy of Server-Side Outputs}]
\label{corollary:c2s_postprocessing}
Suppose the server receives only the privatized latent-state transcript
and no additional unperturbed client information. The learned server
parameters, attention coefficients, Jacobian operators, predictions, and
root-cause scores computed from this transcript satisfy the same
$(\varepsilon_h,\delta_h)$-differential privacy guarantee with respect
to client $m$.
\end{corollary}

\begin{proof}
The result follows from the post-processing property of differential
privacy.
\end{proof}

\begin{proposition}[\textbf{Latent-State Perturbation Error}]
\label{proposition:latent_perturbation}
For one client-to-server release,
\begin{equation}
\mathbb{E}
\left[
\left\|
\widetilde{\mathbf{H}}_m^{(r)}
-
\widehat{\mathbf{H}}_m^{(r)}
\right\|_F^2
\right]
=
p_h\sigma_{h,m}^2.
\end{equation}
\end{proposition}

\begin{proof}
The perturbation is a zero-mean isotropic Gaussian vector with $p_h$
components, each having variance $\sigma_{h,m}^2$.
\end{proof}

Proposition~\ref{proposition:latent_perturbation} shows that the direct
perturbation error grows linearly with the number of communicated latent
variables and quadratically with the noise standard deviation. This
mechanism therefore provides privacy against the server at the cost of
perturbing the state trajectories used to learn the server dynamics.

\subsubsection{Server-to-Client Privacy}

We next consider a \textbf{\textit{trusted server}} and \textbf{\textit{honest-but-curious
clients}}. In this setting, the clients transmit their unperturbed latent
states to the server. The server privatizes the partial gradients sent
back to each client, preventing a receiving client from inferring the
contribution of another client.

Let the server loss during training round $r$ be
$\mathcal{L}_s^{(r)}(D)$. For receiving client $m$, define the partial
gradient
\begin{equation}
G_m^{(r)}(D)
=
\nabla_{\widehat{\mathbf{H}}_m^{(r)}}
L_s^{(r)}(D).
\end{equation}

Because the server model couples the client states, $G_m^{(r)}(D)$ may
depend on the latent states of every client. We therefore analyze its
privacy with respect to a protected client $n\neq m$.

\paragraph{Private Gradient Mechanism.}
The server clips the partial gradient according to
\begin{equation}
\overline{G}_m^{(r)}(D)
=
\operatorname{Clip}
\left(
G_m^{(r)}(D),
C_g
\right)
\end{equation}
and releases
\begin{equation}
\widetilde{G}_m^{(r)}
=
\overline{G}_m^{(r)}(D)
+
Z_{g,m}^{(r)},
\end{equation}
where
\begin{equation}
\operatorname{vec}
\left(
Z_{g,m}^{(r)}
\right)
\sim
\mathcal{N}
\left(
0,
\sigma_g^2 I_{p_g}
\right).
\end{equation}
Here, $p_g$ is the number of scalar elements in the gradient and
$\sigma_g$ is the actual noise standard deviation.

\begin{lemma}[\textbf{Sensitivity of the Clipped Gradient}]
\label{lemma:gradient_sensitivity}
For a fixed receiving client $m$ and protected client $n\neq m$, the
clipped-gradient query
\begin{equation}
f_{m,r}(D)
=
\operatorname{Clip}
\left(
G_m^{(r)}(D),
C_g
\right)
\end{equation}
has global Frobenius sensitivity at most $2C_g$ under replacement
adjacency.
\end{lemma}

\begin{proof}
For any neighboring datasets $D\simeq_n D'$, the triangle inequality
gives
\begin{equation}
\begin{aligned}
&
\left\|
f_{m,r}(D)-f_{m,r}(D')
\right\|_F
\\
&\leq
\left\|
f_{m,r}(D)
\right\|_F
+
\left\|
f_{m,r}(D')
\right\|_F
\\
&\leq
C_g+C_g
=
2C_g.
\end{aligned}
\end{equation}
Therefore, the sensitivity of the clipped-gradient query is at most
$2C_g$.
\end{proof}

\begin{theorem}[\textbf{Server-to-Client Communication}]
\label{theorem:privacy_s2c_single}
Fix a receiving client $m$ and a protected client $n\neq m$. At any
training round $r$, the mechanism by which the server releases
$\widetilde{G}_m^{(r)}$ satisfies
$(\varepsilon_g,\delta_g)$-differential privacy with respect to the
protected data of client $n$, provided that
\begin{equation}
\sigma_g
\geq
\frac{
2C_g\sqrt{2\log(1.25/\delta_g)}
}{
\varepsilon_g
},
\qquad
\varepsilon_g\in(0,1).
\end{equation}
\end{theorem}

\begin{proof}
By Lemma~\ref{lemma:gradient_sensitivity}, the sensitivity of the clipped
gradient is at most $2C_g$. The result follows by applying the Gaussian
mechanism.
\end{proof}

\begin{theorem}[\textbf{Privacy of the Server-to-Client Transcript}]
\label{theorem:privacy_s2c_transcript}
Suppose client $m$ receives $R_g$ independently perturbed gradients,
\begin{equation}
\mathcal{T}_{g,m}
=
\left(
\widetilde{G}_m^{(1)},
\ldots,
\widetilde{G}_m^{(R_g)}
\right),
\end{equation}
using clipping threshold $C_g$ and noise standard deviation $\sigma_g$.
For every $\alpha>1$, the complete transcript satisfies
\begin{equation}
\left(
\alpha,
\frac{
2R_g\alpha C_g^2
}{
\sigma_g^2
}
\right)
\text{-R\'enyi differential privacy}.
\end{equation}
Consequently, for every $\delta_g\in(0,1)$, it satisfies
$(\varepsilon_g,\delta_g)$-differential privacy, where
\begin{equation}
\varepsilon_g
=
\min_{\alpha>1}
\left\{
\frac{
2R_g\alpha C_g^2
}{
\sigma_g^2
}
+
\frac{\log(1/\delta_g)}{\alpha-1}
\right\}.
\end{equation}
\end{theorem}

\begin{proof}
Each gradient release has sensitivity $2C_g$. Therefore, it satisfies
\begin{equation}
\begin{aligned}
\varepsilon_{\alpha}^{(r)}
&=
\frac{
\alpha(2C_g)^2
}{
2\sigma_g^2
}
\\
&=
\frac{
2\alpha C_g^2
}{
\sigma_g^2
}
\end{aligned}
\end{equation}
R\'enyi differential privacy. Composing the $R_g$ releases and converting
the resulting R\'enyi guarantee to
$(\varepsilon_g,\delta_g)$-differential privacy proves the result.
\end{proof}

\begin{corollary}[\textbf{Server-to-Client Noise Calibration}]
\label{corollary:s2c_noise_calibration}
For a target privacy budget $(\varepsilon_g,\delta_g)$ over $R_g$
gradient communications, it is sufficient to choose
\begin{equation}
\sigma_g
\geq
\frac{
C_g\sqrt{2R_g}
}{
\sqrt{\log(1/\delta_g)+\varepsilon_g}
-
\sqrt{\log(1/\delta_g)}
}.
\end{equation}
Equivalently, defining the gradient noise multiplier as
$\nu_g=\sigma_g/C_g$, it is sufficient to choose
\begin{equation}
\nu_g
\geq
\frac{
\sqrt{2R_g}
}{
\sqrt{\log(1/\delta_g)+\varepsilon_g}
-
\sqrt{\log(1/\delta_g)}
}.
\end{equation}
\end{corollary}

\begin{proof}
Let
\begin{equation}
c_g
=
\frac{
2R_gC_g^2
}{
\sigma_g^2
}
\end{equation}
and let $L_g=\log(1/\delta_g)$. Minimizing the expression in
Theorem~\ref{theorem:privacy_s2c_transcript} over $\alpha>1$ gives
\begin{equation}
\varepsilon_g
=
c_g+2\sqrt{c_gL_g}.
\end{equation}
Solving for $\sigma_g$ gives the stated result.
\end{proof}

\begin{corollary}[\textbf{Privacy of Client Updates}]
\label{corollary:s2c_postprocessing}
Any client-model update computed exclusively from the privatized
gradient transcript $\mathcal{T}_{g,m}$ and the receiving client's local
data satisfies the same $(\varepsilon_g,\delta_g)$-differential privacy
guarantee with respect to the protected client.
\end{corollary}

\begin{proof}
The result follows from the post-processing property of differential
privacy.
\end{proof}

\begin{proposition}[\textbf{Gradient Perturbation Error}]
\label{proposition:gradient_perturbation}
For one server-to-client communication,
\begin{equation}
\begin{aligned}
&
\mathbb{E}
\left[
\left\|
\widetilde{G}_m^{(r)}
-
G_m^{(r)}
\right\|_F^2
\right]
\\
&=
\left\|
G_m^{(r)}
-
\operatorname{Clip}
\left(
G_m^{(r)},
C_g
\right)
\right\|_F^2
+
p_g\sigma_g^2.
\end{aligned}
\end{equation}
\end{proposition}

\begin{proof}
The perturbation can be decomposed as
\begin{equation}
\widetilde{G}_m^{(r)}-G_m^{(r)}
=
\operatorname{Clip}
\left(
G_m^{(r)},
C_g
\right)
-
G_m^{(r)}
+
Z_{g,m}^{(r)}.
\end{equation}
The Gaussian noise is zero-mean and independent of the unperturbed
gradient, so the expected cross term is zero. Furthermore,
\begin{equation}
\mathbb{E}
\left[
\left\|
Z_{g,m}^{(r)}
\right\|_F^2
\right]
=
p_g\sigma_g^2.
\end{equation}
Combining these expressions proves the result.
\end{proof}

Proposition~\ref{proposition:gradient_perturbation} separates the
gradient distortion into the bias introduced by clipping and the
variance introduced by Gaussian perturbation. Unlike client-to-server
state perturbation, this mechanism leaves unchanged the latent
trajectories used by the server to learn the system dynamics. It instead
perturbs the learning signal returned to the clients.

\subsection{Privacy Evaluation}
\label{appendix:evaluation}

We evaluate the two communication directions separately. For \textit{client-to-server} privacy, we fix $\delta_h$, the sensitivity bound
$\Delta_{h,m}$, and the number of latent-state releases $R_h$. For each
privacy budget $\varepsilon_h$, we compute $\sigma_{h,m}$ using
Corollary~\ref{corollary:c2s_noise_calibration}, perturb the communicated
latent states, and train the complete client-server framework. For \textit{server-to-client} privacy, we fix $\delta_g$, the clipping threshold
$C_g$, and the number of gradient releases $R_g$. For each privacy budget
$\varepsilon_g$, we compute $\sigma_g$ using
Corollary~\ref{corollary:s2c_noise_calibration}, clip and perturb the
partial gradients, and train the complete framework.

For both settings, we report two interdependency recovery metrics -- \textbf{(1)} the server loss $L_s$ and \textbf{(2)} the cosine
similarity between the learned and ground-truth Jacobian operators as
functions of $\varepsilon$. Smaller values of $\varepsilon$ correspond
to stronger privacy. The non-private framework is included as a
reference. For server-to-client privacy, we additionally include a
clipping-only baseline to separate the effects of clipping from those of
Gaussian perturbation. The quantities $R_h$ and $R_g$ represent the numbers of newly released
data-dependent messages. Reusing a previously privatized message does
not incur additional privacy loss. The analysis does not use privacy
amplification through random subsampling and is therefore conservative
when communication rounds operate on randomly sampled temporal windows.

\paragraph{\textbf{Setup.}}
The privacy--utility experiments use the $M{=}6$ nonlinear SSM
configuration of the synthetic evaluation
(Tables~\ref{tab:prediction} and \ref{tab:structure}), so that the
private runs are directly comparable to the non-private results
reported there. For every privacy configuration the framework is
retrained from scratch with the train--test split, architectures,
optimizer, epoch budget, and calibration constants held fixed. All
reported curves show the mean over three random seeds, and the shaded
bands show $\pm 1$ standard deviation across those seeds. We use
the privacy grid $\varepsilon\in\{0.5,1,2,4,8\}$ together with the
non-private reference $\varepsilon=\infty$ and, for the client-to-server
direction, extension points $\varepsilon\in\{16,64,256\}$ at
$\delta=10^{-5}$; $\delta\in\{10^{-3},10^{-5},10^{-7}\}$ are shown as
separate curves.

\begin{figure}[htbp]
\centering
\includegraphics[width=0.49\columnwidth]{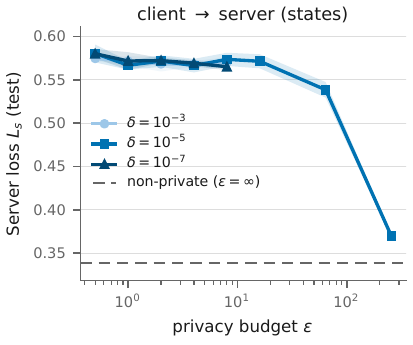}\hfill
\includegraphics[width=0.49\columnwidth]{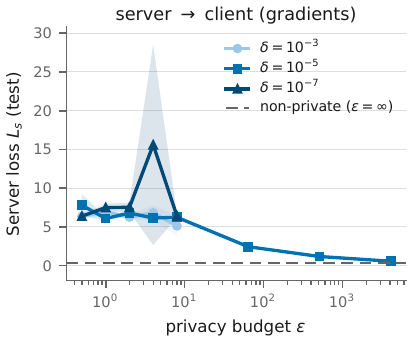}
\caption{Server loss on the test window versus $\varepsilon$ for
the two communication directions (one curve per $\delta$), normalized
per client and state dimension. Left: client-to-server latent-state
perturbation. Right: server-to-client gradient perturbation. Dashed:
non-private reference. Shaded bands: $\pm 1$ standard deviation over
seeds.}
\label{fig:dp-tradeoff-sls}
\end{figure}

\begin{figure}[htbp]
\centering
\includegraphics[width=0.49\columnwidth]{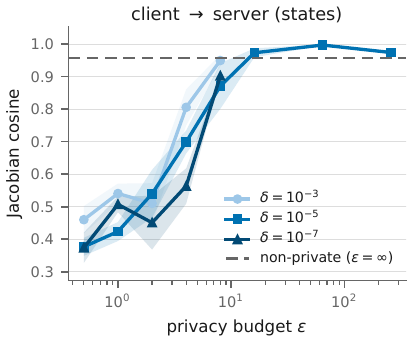}\hfill
\includegraphics[width=0.49\columnwidth]{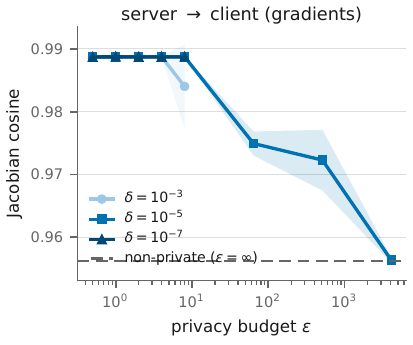}
\caption{Interpretability under privacy on the $M{=}6$ nonlinear SSM:
cosine between the learned interdependency structure and the
ground-truth interaction structure.}
\label{fig:dp-tradeoff-cos}
\end{figure}

\paragraph{\textbf{Results.}}
For client-to-server perturbation, Figure~\ref{fig:dp-tradeoff-sls}
shows that the normalized server loss increases from \(0.34\) in the
non-private setting to approximately \(0.57\) for
\(\varepsilon\leq16\), corresponding to a \(68\%\) increase. The loss
then decreases to \(0.54\) at \(\varepsilon=64\) and \(0.37\) at
\(\varepsilon=256\), which is within \(9\%\) of the non-private
reference. Figure~\ref{fig:dp-tradeoff-cos} shows a corresponding loss
of interpretability under strict privacy. The Jacobian cosine decreases
from \(0.956\) to \(0.37\text{--}0.54\) for
\(\varepsilon\leq2\), a \(44\%\text{--}61\%\) reduction. It recovers to
\(0.70\) at \(\varepsilon=4\), \(0.87\) at \(\varepsilon=8\), and at
least \(0.97\) for \(\varepsilon\geq16\). Thus, perturbing the
communicated states affects both server fit and interdependency
recovery, while both recover as the privacy budget increases.

For server-to-client perturbation, Figure~\ref{fig:dp-tradeoff-sls}
shows that the server loss decreases monotonically as the gradient noise
vanishes. Relative to the non-private loss of \(0.34\), the loss is
\(2.4\) at \(\varepsilon=64\), \(1.2\) at
\(\varepsilon=512\), and \(0.56\) at
\(\varepsilon=4096\), corresponding to increases of approximately
\(606\%\), \(253\%\), and \(65\%\), respectively. In contrast,
Figure~\ref{fig:dp-tradeoff-cos} shows that the Jacobian cosine remains
\(0.989\) throughout the standard privacy grid and between \(0.96\) and
\(0.99\) over the extended budgets, compared with the non-private
reference of \(0.956\). The server continues to learn from the
unperturbed latent states, which preserves the interdependency
structure. However, noisy gradient feedback degrades the client updates
and therefore increases the discrepancy between the server predictions
and client targets.

\section{Algorithm Pseudocode}
\label{app:algorithms}
Algorithm~\ref{alg:client-model} details the proprietary model at the clients
whose states enter the federated training procedure, summarized in
Algorithm~\ref{alg:fed-harmonized}, as inputs. The local EKF realization in
Algorithm~\ref{alg:client-model} is the instantiation used in our
synthetic experiments. The framework treats the proprietary model as a
black box that supplies the local state estimates $(\tilde h_t^m)_c$,
$(\hat h_t^m)_c$: any model that remains proprietary can take its place;
in the real-world experiments, the clients' proprietary models are
pretrained LSTMs.

\begin{algorithm}[htbp]
\caption{Proprietary Model at the Client}
\label{alg:client-model}
\begin{algorithmic}[1]
\State \textbf{Inputs}: $T$, $f_m$, $g_m$
\State \textbf{Initialize at Client} $m$: $(\hat h_0^m)_c$, $P_0^m$
\For{$t = 1$ \textbf{to} $T$}
    \State $(\tilde h_t^m)_c \gets f_m\big((\hat h_{t-1}^m)_c\big)$
    \State $(r_t^m)_c \gets y_t^m - g_m\big((\tilde h_t^m)_c\big)$
    \State $(\hat h_t^m)_c \gets (\tilde h_t^m)_c + K_t^m \cdot (r_t^m)_c$
    \State Update $K_t^m$, $P_t^m$ via the EKF recursion
\EndFor
\end{algorithmic}
\end{algorithm}

\begin{algorithm}[htbp]
\caption{Federated Learning of Nonlinear Cross-Client Temporal Interdependencies}
\label{alg:fed-harmonized}
\begin{algorithmic}[1]
\State \textbf{Inputs}: $T$, $f_m$, $g_m$, $\mathcal{N}(\cdot)$,
$\{(\hat h_t^m)_c\}\ \forall m \in \{1,\dots,M\},\, t \in \{1,\dots,T\}$
\State \textbf{Choose}: \textit{epoch}, \textit{tol}, $k = 0$
\State \textbf{Initialize at Server}: GAT parameters (yielding $\{\alpha_{mn}, W_{mn}\}$ over $\mathcal{N}(\cdot)$)
\State \textbf{Choose at Server}: $\gamma$
\State \textbf{Initialize at Client} $m$: $\theta_m^0$
\State \textbf{Choose at Client}: $\eta_1,\ \eta_2$
\While{$k < \textit{epoch} \cdot T$ \textbf{and} $L_s > \textit{tol}$}
    \For{$t = 1$ \textbf{to} $T$}
        \For{each client $m$}
            \State $(\hat h_{t-1}^m)_a \gets (\hat h_{t-1}^m)_c + \Delta_m\big(y_{t-1}^m;\,\theta_m^k\big)$
            \State $(\tilde h_t^m)_a \gets f_m\big((\hat h_{t-1}^m)_a\big)$
            \State Send $\big[(\tilde h_t^m)_a,\ (\hat h_{t-1}^m)_c\big]$ to the server
        \EndFor
        \State \textbf{At the server}:
        \Statex \hspace{2.6em} $(\tilde H_t)_a \gets \big[(\tilde h_t^1)_a, \dots, (\tilde h_t^M)_a\big]^{\top}$
        \Statex \hspace{2.6em} $(\tilde H_t)_s \gets \big[(\tilde h_t^1)_s, \dots, (\tilde h_t^M)_s\big]^{\top}$ where, 
        $(\tilde h_t^m)_s = \sigma\Big(\sum\limits_{n \in \mathcal{N}(m)} \alpha_{mn}\, W_{mn}\, (\hat h_{t-1}^n)_c\Big)$
        \Statex \hspace{2.6em} $L_s \gets \big\| (\tilde H_t)_a - (\tilde H_t)_s \big\|_2^2$
        \Statex \hspace{2.6em} Send $\nabla_{(\tilde h_t^m)_a} L_s$ to the client $m$
        \Statex \hspace{2.6em} GAT parameters $\gets$ GAT parameters $- \gamma \cdot \nabla L_s$
        \For{each client $m$}
            \State $\theta_m^{k+1} \gets \theta_m^{k} - \eta_1 \cdot \nabla_{\theta_m^k} (L_m)_a
            - \eta_2 \cdot \nabla_{\theta_m^k} L_s$
        \EndFor
        \State $k \gets k+1$
    \EndFor
\EndWhile
\State \textbf{Output}: at the server, the trained GAT with its interdependency
read-outs $\alpha_{mn}(t)$ and $J_{mn}(t)$; at each client $m$, the trained
augmentation parameters $\theta_m$
\end{algorithmic}
\end{algorithm}

\section{Proofs}
\label{app:proofs}
\subsection{Proof of Proposition~\ref{prop:jacobian_gat}}
Recall the server-side GAT transition at client $m$:
\[
(\tilde h_t^m)_s
=
\sigma\!\Big(s_m(t)\Big),
\hspace{0.1cm} \text{with} \hspace{0.1cm}
s_m(t):=
\sum_{r\in\mathcal N(m)} \alpha_{mr}(t)\, W_{mr}\,(\hat h_{t-1}^r)_c .
\]
Fix $m,n,t$ and differentiate $(\tilde h_t^m)_s$ w.r.t.\ $(\hat h_{t-1}^n)_c$.
By the chain rule,
\begin{equation}
\label{eq:chainrule_J}
J_{mn}(t)
:=
\frac{\partial (\tilde h_t^m)_s}{\partial (\hat h_{t-1}^n)_c}
=
\operatorname{diag}\!\big(\sigma'(s_m(t))\big)\;
\frac{\partial s_m(t)}{\partial (\hat h_{t-1}^n)_c}.
\end{equation}
It remains to compute $\frac{\partial s_m(t)}{\partial (\hat h_{t-1}^n)_c}$.
Using the product rule on each summand of $s_m(t)$,
\[
\frac{\partial s_m(t)}{\partial (\hat h_{t-1}^n)_c}
=
\sum_{r\in\mathcal N(m)}
\Bigg[
\frac{\partial \alpha_{mr}(t)}{\partial (\hat h_{t-1}^n)_c}\; W_{mr}\,(\hat h_{t-1}^r)_c
\;+\;
\alpha_{mr}(t)\,W_{mr}\,\frac{\partial (\hat h_{t-1}^r)_c}{\partial (\hat h_{t-1}^n)_c}
\Bigg].
\]
Since $\frac{\partial (\hat h_{t-1}^r)_c}{\partial (\hat h_{t-1}^n)_c} = \delta_{rn} I$,
the second term yields the direct pathway $\alpha_{mn}(t)W_{mn}$.
For the first term, recall that the attention coefficients are softmax-normalized:
\[
\alpha_{mr}(t)=\frac{\exp(e_{mr}(t))}{\sum_{q\in\mathcal N(m)}\exp(e_{mq}(t))}.
\]
The standard softmax derivative gives, for any $r\in\mathcal N(m)$,
\[
\frac{\partial \alpha_{mr}(t)}{\partial e_{mn}(t)}
=
\alpha_{mr}(t)\big(\delta_{rn}-\alpha_{mn}(t)\big).
\]
By the chain rule,
\[
\frac{\partial \alpha_{mr}(t)}{\partial (\hat h_{t-1}^n)_c}
=
\frac{\partial \alpha_{mr}(t)}{\partial e_{mn}(t)}\;
\frac{\partial e_{mn}(t)}{\partial (\hat h_{t-1}^n)_c}
=
\alpha_{mr}(t)\big(\delta_{rn}-\alpha_{mn}(t)\big)\;
\frac{\partial e_{mn}(t)}{\partial (\hat h_{t-1}^n)_c}.
\]
Substituting into the first term gives
\[
\sum_{r\in\mathcal N(m)}
W_{mr}\,(\hat h_{t-1}^r)_c\;
\alpha_{mr}(t)\big(\delta_{rn}-\alpha_{mn}(t)\big)\;
\frac{\partial e_{mn}(t)}{\partial (\hat h_{t-1}^n)_c}.
\]
Combining the direct and indirect contributions, we obtain
\begin{align*}
\frac{\partial s_m(t)}{\partial (\hat h_{t-1}^n)_c}
=
\alpha_{mn}(t)\, W_{mn}
&+
\sum_{r \in \mathcal{N}(m)}
W_{mr}\,(\hat h_{t-1}^r)_c\;
\alpha_{mr}(t)\\& \times \big(\delta_{rn} - \alpha_{mn}(t)\big)
\frac{\partial e_{mn}(t)}{\partial (\hat h_{t-1}^n)_c}.
\end{align*}
Finally, substituting this expression into \eqref{eq:chainrule_J} yields
Eq.~\eqref{eq:jacobian_alpha}, proving the proposition.
\qed


\subsection{Proof of Lemma~\ref{lemma:stateboundedness}}
\textbf{Assumption (Stable EKF).}
For each client $m$, assume the oracle EKF and the proprietary client EKF are
mean-square stable in the following sense: there exist constants
$0<\kappa<1$ and $B_o,B_c<\infty$ such that the corresponding estimation errors
satisfy, for all $t$,
\begin{align}
\mathbb{E}\big\|(\hat h_o^{t})_m\big\| \le B_o,
\qquad
\mathbb{E}\big\|(\hat h_c^{t})_m\big\| \le B_c,
\label{eq:ekf-bounded}
\end{align}
and the one-step oracle prediction map is Lipschitz on the operating region.
Such bounds follow from standard EKF stability results under bounded
process/observation noise and local Lipschitz dynamics.
\\
\\
\textbf{(i)} Define $E_t^m := \mathbb{E}\|(\hat h_c^{t})_m-(\hat h_o^{t})_m\|$.
Using the EKF update forms for the client and oracle and Lipschitz continuity
of the observation function, we obtain
a standard recursion of the form
\begin{equation}
E_t^m \;\le\; a_m \;+\; b_m\,E_{t-1}^m,
\qquad\text{for some } a_m<\infty,\; 0<b_m<1,
\label{eq:Et-rec}
\end{equation}
where $a_m$ aggregates bounded residual/noise terms,
and $b_m$ is a contraction constant determined by the local Lipschitz factors.
Iterating Eq.~\eqref{eq:Et-rec} yields
\[
\limsup_{t\to\infty} E_t^m \;\le\; \frac{a_m}{1-b_m} \;=:\; \bar\varepsilon_2,
\]
which proves the existence of $\bar\varepsilon_2\ge 0$ such that
$\|(\hat h_c^t)_m-(\hat h_o^t)_m\|\le \bar\varepsilon_2$ for sufficiently large $t$
(in expectation). For the transformed neighbor representations of
Eq.~\eqref{eq:transformed}, adding and subtracting
$W_{mn}^{s}(\hat h_o^{t-1})_n$ gives
\[
\|z_{mn}^{s}-z_{mn}^{o}\|
\le \|W_{mn}^{s}\|\,\|(\hat h_c^{t-1})_n-(\hat h_o^{t-1})_n\|
+ \|W_{mn}^{s}-W_{mn}^{o}\|\,\|(\hat h_o^{t-1})_n\| .
\]
Since the transformations are finite and the oracle states are bounded by
\eqref{eq:ekf-bounded}, there exists $\varepsilon_2\ge 0$ such that
$\|z_{mn}^{s}-z_{mn}^{o}\|\le\varepsilon_2$ for all $n$ and sufficiently
large $t$. Stacking the columns yields
$\|Z_{s,m}^{t-1}-Z_{o,m}^{t-1}\|_F\le\sqrt{N_m}\,\varepsilon_2$.
\\
\\
\textbf{(ii)}
Recall the GAT equations from the paper,
\[
(\tilde h_o^t)_m=\sigma(Z_{o,m}^{t-1}\alpha_m^o),
\qquad
(\tilde h_s^t)_m=\sigma(Z_{s,m}^{t-1}\alpha_m^s).
\]
Let $\Delta_t^m := \mathbb{E}\|(\tilde h_s^t)_m-(\tilde h_o^t)_m\|$.
By Lipschitz continuity of $\sigma$ and boundedness of
states \eqref{eq:ekf-bounded}, we obtain,
\begin{align*}
\Delta_t^m
&\le L_\sigma\,\mathbb{E}\|Z_{s,m}^{t-1}\alpha_m^s - Z_{o,m}^{t-1}\alpha_m^o\| \\
&\le L_\sigma\Big(
\mathbb{E}\|(Z_{s,m}^{t-1}-Z_{o,m}^{t-1})\alpha_m^o\|
+
\mathbb{E}\|Z_{s,m}^{t-1}(\alpha_m^s-\alpha_m^o)\|
\Big).
\end{align*}
The first term is bounded by part (i) (each column difference of
$Z_{s,m}^{t-1}-Z_{o,m}^{t-1}$ is at most $\varepsilon_2$), and the
second term is bounded because $Z_{s,m}^{t-1}$ is bounded on the operating set and
$\alpha_m^s,\alpha_m^o$ are bounded by construction of attention (softmax).
Therefore there exists $\varepsilon_1\ge 0$ such that
\[
\limsup_{t\to\infty}\Delta_t^m \;\le\; \varepsilon_1,
\]
i.e.,
$\|(\tilde h_s^t)_m-(\tilde h_o^t)_m\|\le \varepsilon_1$ for sufficiently large
$t$ (in expectation). This completes the proof.
\qed

\subsection{Proof of Theorem~\ref{thm:param_conv}}
Fix a client (node) $m$ with $N_m$ neighbors and suppress $m$ from notation where
clear. From \eqref{eq:update} and the definitions of $Z_{o,m}^{t-1},Z_{s,m}^{t-1}$ and
$\alpha_m^o,\alpha_m^s$, we can write
\[
(\tilde h_o^t)_m = \sigma\!\big(Z_{o,m}^{t-1}\alpha_m^o\big),
\qquad
(\tilde h_s^t)_m = \sigma\!\big(Z_{s,m}^{t-1}\alpha_m^s\big).
\]
By Lemma~\ref{lemma:stateboundedness},
\[
\big\|\sigma(Z_{s,m}^{t-1}\alpha_m^s)-\sigma(Z_{o,m}^{t-1}\alpha_m^o)\big\|
=
\|(\tilde h_s^t)_m-(\tilde h_o^t)_m\|
\le \varepsilon_1.
\]
Using the bi-Lipschitz inverse bound of $\sigma$ (Assumption: Bi-Lipschitz),
\begin{equation}
\label{eq:preact_align}
\big\|Z_{s,m}^{t-1}\alpha_m^s - Z_{o,m}^{t-1}\alpha_m^o\big\|
\le L_{\sigma^{-1}}\,\varepsilon_1.
\end{equation}
Add and subtract $Z_{s,m}^{t-1}\alpha_m^o$ and apply the triangle inequality:
\[
\big\|Z_{s,m}^{t-1}\alpha_m^s - Z_{o,m}^{t-1}\alpha_m^o\big\|
\ge
\big\|Z_{s,m}^{t-1}(\alpha_m^s-\alpha_m^o)\big\|
-
\big\|(Z_{s,m}^{t-1}-Z_{o,m}^{t-1})\alpha_m^o\big\|.
\]
Rearranging and combining with \eqref{eq:preact_align} yields
\begin{equation}
\label{eq:Hc_deltaalpha_bound}
\big\|Z_{s,m}^{t-1}(\alpha_m^o-\alpha_m^s)\big\|
\le
\big\|(Z_{s,m}^{t-1}-Z_{o,m}^{t-1})\alpha_m^o\big\|
+
L_{\sigma^{-1}}\,\varepsilon_1.
\end{equation}
Next, by Lemma~\ref{lemma:stateboundedness}, each neighbor column satisfies
$\|z_{mn}^{s}-z_{mn}^{o}\|\le \varepsilon_2$, hence there exists a
constant $C_h>0$ (depending only on $\|\alpha_m^o\|$ and $N_m$) such that
\begin{equation}
\label{eq:Hdiff_alpha_bound}
\big\|(Z_{s,m}^{t-1}-Z_{o,m}^{t-1})\alpha_m^o\big\|\le C_h\,\varepsilon_2.
\end{equation}
Substituting \eqref{eq:Hdiff_alpha_bound} into \eqref{eq:Hc_deltaalpha_bound} gives
\begin{equation}
\label{eq:Hc_deltaalpha_bound2}
\big\|Z_{s,m}^{t-1}(\alpha_m^o-\alpha_m^s)\big\|
\le
C_h\,\varepsilon_2 + L_{\sigma^{-1}}\,\varepsilon_1.
\end{equation}
Finally, decompose the attention discrepancy along the row space and the
null space of $Z_{s,m}^{t-1}$,
\[
\alpha_m^o-\alpha_m^s
=
P_{\operatorname{row}(Z_{s,m}^{t-1})}(\alpha_m^o-\alpha_m^s)
+
P_{\ker(Z_{s,m}^{t-1})}(\alpha_m^o-\alpha_m^s),
\]
where $P_{\operatorname{row}(Z_{s,m}^{t-1})}$ and $P_{\ker(Z_{s,m}^{t-1})}$ denote the
orthogonal projections onto the row space and the null space of $Z_{s,m}^{t-1}$.
Since $Z_{s,m}^{t-1}P_{\ker(Z_{s,m}^{t-1})}(\alpha_m^o-\alpha_m^s)=0$, we have
$Z_{s,m}^{t-1}(\alpha_m^o-\alpha_m^s)
=Z_{s,m}^{t-1}P_{\operatorname{row}(Z_{s,m}^{t-1})}(\alpha_m^o-\alpha_m^s)$.
For any vector $x$ in the row space of $Z_{s,m}^{t-1}$,
$\|Z_{s,m}^{t-1}x\|\ge \sigma_{\min}^{+}(Z_{s,m}^{t-1})\|x\|$, where
$\sigma_{\min}^{+}(Z_{s,m}^{t-1})$ denotes the smallest nonzero singular value of
$Z_{s,m}^{t-1}$; no rank assumption on $Z_{s,m}^{t-1}$ is needed. Applying this with
$x=P_{\operatorname{row}(Z_{s,m}^{t-1})}(\alpha_m^o-\alpha_m^s)$ and using
\eqref{eq:Hc_deltaalpha_bound2},
\[
\big\|P_{\operatorname{row}(Z_{s,m}^{t-1})}(\alpha_m^o - \alpha_m^s)\big\|
\;\le\;
\frac{C_h\,\varepsilon_2 + L_{\sigma^{-1}}\,\varepsilon_1}
     {\sigma_{\min}^{+}(Z_{s,m}^{t-1})}.
\]
This proves Theorem~\ref{thm:param_conv}.
\qed

\subsection{Proof of Theorem~\ref{thm:J_conv}}
Fix a client $m$ and a neighbor $n\in\mathcal N(m)$. Recall the Jacobian mapping
$
\mathcal J_{mn}:(\alpha_m,Z)\mapsto J_{mn}(\alpha_m,Z),
$
with
\[
J^{(s)}_{mn}(t)=\mathcal J_{mn}(\alpha_m^s,Z_{s,m}^{t-1}),
\qquad
J^{(o)}_{mn}(t)=\mathcal J_{mn}(\alpha_m^o,Z_{o,m}^{t-1}).
\]
By the Lipschitz continuity assumption on $\mathcal J_{mn}$, we have
\begin{equation}
\label{eq:Jlip_apply}
\big\|J^{(s)}_{mn}(t)-J^{(o)}_{mn}(t)\big\|
\le
L_J\Big(\|\alpha_m^s-\alpha_m^o\|+\|Z_{s,m}^{t-1}-Z_{o,m}^{t-1}\|_F\Big).
\end{equation}
By Lemma~\ref{lemma:stateboundedness} we have,
\begin{equation}
\label{eq:Hfro_bound}
\|Z_{s,m}^{t-1}-Z_{o,m}^{t-1}\|_F
=
\Big(\sum_{j=1}^{N_m}\|z_{mj}^{s}-z_{mj}^{o}\|^2\Big)^{1/2}
\le \sqrt{N_m}\,\varepsilon_2.
\end{equation}
Moreover, since the row-space and null-space components of
$\alpha_m^o-\alpha_m^s$ with respect to $Z_{s,m}^{t-1}$ are orthogonal,
\begin{equation}
\label{eq:alpha_decomp}
\|\alpha_m^o-\alpha_m^s\|
\le
\big\|P_{\operatorname{row}(Z_{s,m}^{t-1})}(\alpha_m^o-\alpha_m^s)\big\|
+
\big\|P_{\ker(Z_{s,m}^{t-1})}(\alpha_m^o-\alpha_m^s)\big\|,
\end{equation}
and Theorem~\ref{thm:param_conv} bounds the first term,
\begin{equation}
\label{eq:alpha_bound_use}
\big\|P_{\operatorname{row}(Z_{s,m}^{t-1})}(\alpha_m^o-\alpha_m^s)\big\|
\le
\frac{C_h\,\varepsilon_2 + L_{\sigma^{-1}}\,\varepsilon_1}
     {\sigma_{\min}^{+}(Z_{s,m}^{t-1})}.
\end{equation}
Substituting \eqref{eq:Hfro_bound}, \eqref{eq:alpha_decomp}, and
\eqref{eq:alpha_bound_use} into \eqref{eq:Jlip_apply} yields
\begin{multline*}
\big\|J^{(s)}_{mn}(t) - J^{(o)}_{mn}(t)\big\|
\le
L_J\Bigg(
\frac{C_h\,\varepsilon_2 + L_{\sigma^{-1}}\,\varepsilon_1}
     {\sigma_{\min}^{+}(Z_{s,m}^{t-1})}
\\
+ \big\|P_{\ker(Z_{s,m}^{t-1})}(\alpha_m^o-\alpha_m^s)\big\|
+ \sqrt{N_m}\,\varepsilon_2
\Bigg),
\end{multline*}
which proves Theorem~\ref{thm:J_conv}.
\qed

\end{document}